\documentclass[10pt,journal,compsoc]{IEEEtran}

\ifCLASSOPTIONcompsoc
  \usepackage[nocompress]{cite}
\else
  \usepackage{cite}
\fi
\ifCLASSINFOpdf
\else
\fi
\usepackage{amsmath}
\usepackage{amssymb}
\usepackage{mathtools}
\usepackage{amsthm}

\usepackage{amsfonts}
\usepackage{multirow}
\usepackage{multicol}

\usepackage{booktabs}
\usepackage{color}
\usepackage{algorithm}
\usepackage{algorithmic}

\usepackage{graphicx}
\usepackage{subfig}

\newtheorem{theorem}{Theorem}
\newtheorem{lemma}{Lemma}

\newtheorem{definition}{Definition}
\newtheorem{assumption}{Assumption}
\newtheorem{remark}{Remark}

\begin{document}
%
\title{Gradient Descent Ascent for Minimax Problems on Riemannian Manifolds}
%
%
%
%

\author{Feihu~Huang,~Shangqian~Gao 
\IEEEcompsocitemizethanks{\IEEEcompsocthanksitem Feihu Huang is with College of Computer Science and Technology, Nanjing University of Aeronautics and Astronautics, Nanjing, China; and also with MIIT Key Laboratory of Pattern Analysis and Machine Intelligence, Nanjing, China. 
E-mail: huangfeihu2018@gmail.com; huangfeihu@nuaa.edu.cn \\
Shangqian Gao 
is with Department of Electrical and Computer Engineering, University of Pittsburgh, Pittsburgh, USA.
E-mail: shg84@pitt.edu
}}

%
%

\markboth{Journal of \LaTeX\ Class Files,~Vol.~14, No.~8, August~2015}%
{Shell \MakeLowercase{\textit{et al.}}: Bare Advanced Demo of IEEEtran.cls for IEEE Computer Society Journals}
%



\IEEEtitleabstractindextext{%
\begin{abstract}
In the paper, we study a class of useful minimax problems on Riemanian manifolds and propose a class of
effective Riemanian gradient-based methods to solve these minimax problems. Specifically, we propose an effective
Riemannian gradient descent ascent (RGDA) algorithm for the deterministic minimax optimization.
Moreover, we prove that our RGDA has a sample complexity of $O(\kappa^2\epsilon^{-2})$
for finding an $\epsilon$-stationary solution of the Geodesically-Nonconvex Strongly-Concave (GNSC) minimax problems, where $\kappa$ denotes the condition number.
At the same time, we present an effective Riemannian stochastic gradient descent ascent (RSGDA) algorithm for the stochastic minimax optimization, which has a sample complexity of $O(\kappa^4\epsilon^{-4})$ for finding an $\epsilon$-stationary solution. 
To further reduce the sample complexity, we propose an accelerated Riemannian stochastic gradient descent ascent (Acc-RSGDA) algorithm
based on the momentum-based  variance-reduced technique.
We prove that our Acc-RSGDA algorithm achieves a lower sample complexity of $\tilde{O}(\kappa^{4}\epsilon^{-3})$ in searching for an $\epsilon$-stationary solution of the GNSC minimax problems.
Extensive experimental results on the robust distributional optimization and robust
Deep Neural Networks (DNNs) training over Stiefel manifold
demonstrate efficiency of our algorithms.
\end{abstract}

}

\maketitle

\IEEEdisplaynontitleabstractindextext

%
\IEEEpeerreviewmaketitle

\section{Introduction}
\IEEEPARstart{I}{n} this paper, we study a class of useful minimax optimization problems on the Riemannian manifold $\mathcal{M}$, defined as:
\begin{align} \label{eq:1}
 \min_{x \in \mathcal{M}}\max_{y\in \mathcal{Y}}f(x,y),
\end{align}
where function $f(x,y):\mathcal{M} \times \mathcal{Y} \rightarrow \mathbb{R}$ is $\mu$-strongly concave in $y \in \mathcal{Y}\subseteq \mathbb{R}^d$ but possibly (geodesically) nonconvex in $x \in \mathcal{M}$.
Here $\mathcal{M}$ is a Riemannian manifold, and $\mathcal{Y}$ is a
convex and closed set in Euclidean space. $f(\cdot,y): \mathcal{M} \rightarrow \mathbb{R}$ for any $y \in \mathcal{Y}$ is a smooth but possibly (geodesically)  nonconvex
real-valued function on manifold $\mathcal{M}$, and $f(x,\cdot): \mathcal{Y} \rightarrow \mathbb{R}$ for any $x\in \mathcal{M}$ is a smooth and strongly-concave real-valued function. Note that a geodesically nonconvex  function on Riemannian manifold also is nonconvex on Euclidean space, and a geodesically convex  function on Riemannian manifold may be nonconvex on Euclidean space.
In this paper, we also focus on the stochastic form of minimax problem \eqref{eq:1}, defined as
\begin{align} \label{eq:2}
 \min_{x \in \mathcal{M}}\max_{y\in \mathcal{Y}} \mathbb{E}_{\xi\sim\mathcal{D}}[f(x,y;\xi)],
\end{align}
where $\xi$ is a random variable that follows an unknown distribution $\mathcal{D}$.
In fact, Problems (\ref{eq:1}) and (\ref{eq:2}) are associated to many existing machine learning applications:

\textbf{1). Robust DNNs Training over Riemannian manifold.}
Deep Neural Networks (DNNs) recently have been demonstrating exceptional performance on many machine learning applications such as
image classification. However, they are vulnerable to the adversarial example attacks,
which show that a small perturbation in the data input can
significantly change the output of DNNs. Thus, the security properties of DNNs have been widely studied.
One of secured DNN research topics is to enhance the robustness of DNNs under the adversarial example attacks.
Given the training sample $\mathcal{D}:=\{\xi_i=(a_i,b_i)\}_{i=1}^n$,
where $a_i\in \mathbb{R}^d$ and $b_i\in \mathbb{R}$ represent the features and label of sample $\xi_i$ respectively.
Then we train a robust DNN against a universal adversarial attack  \cite{moosavi2017universal,akhtar2018defense}, which can be formulated the following minimax problem:
\begin{align} \label{eq:3}
\min_{x \in \mathbb{R}^q}\max_{y\in \mathcal{Y}}\frac{1}{n}\sum_{i=1}^n\ell(h(a_i+y;x),b_i),
\end{align}
where $x\in \mathbb{R}^q$ denotes weight of the DNN, and $h(\cdot;x)$ denotes
the DNN parameterized by $x$, and $\ell(\cdot)$ is the
loss function.
Here $y$ denotes a small universal perturbation in the features $\{a_i\}_{i=1}^n$,
and the constraint $\mathcal{Y}=\{y: \|y\|_\infty \leq \varepsilon\}$ indicates that the
poisoned samples should not be too different from the original ones.

Recently, the orthonormality on weights of DNNs has gained much interest and has been found to
be useful across different tasks such as person re-identification~\cite{sun2017svdnet} and image classification~\cite{xie2017all}.
In fact, the orthonormality constraints improve the performances of DNNs \cite{li2020efficient,bansal2018can},
and reduce overfitting to improve
generalization \cite{cogswell2015reducing}. At the same time,
the orthonormality can stabilize the distribution of activation over layers within DNNs
~\cite{huang2018orthogonal}. Thus, we further consider the
following robust DNN training over the Stiefel manifold $\mathcal{M}$:
\begin{align}\label{eq:4}
\min_{x \in \mathcal{M}}\max_{y\in \mathcal{Y}}\frac{1}{n}\sum_{i=1}^n\ell(h(a_i+y;x),b_i),
\end{align}
When data are continuously coming, we can rewrite the stochastic form of  Problem (\ref{eq:4}) as follows:
\begin{align} \label{eq:5}
\min_{x \in \mathcal{M}}\max_{y\in \mathcal{Y}} \mathbb{E}_{\xi}[f(x,y;\xi)],
\end{align}
where $f(x,y;\xi)=\ell(h(a+y;x),b)$ with $\xi=(a,b)$.

\textbf{2). Distributionally Robust Optimization over Riemannian manifold. }
Distributionally Robust Optimization (DRO) \cite{chen2017robust,rahimian2019distributionally} is an effective method to
deal with the noisy data, adversarial data, and imbalanced data.
In the paper, we consider the DRO over the Riemannian manifold that can be applied in many machine learning problems such as
robust principal component analysis (PCA) and distributionally robust DNN training.
To be more specific, given a set of data samples $\{\xi_i\}_{i=1}^n$,
the DRO over Riemannian manifold $\mathcal{M}$ can be written
as the following minimax problem:
\begin{align}~\label{eq:6}
 \min_{x\in \mathcal{M}} \max_{\textbf{p}\in \mathcal{S}} \bigg\{ \sum_{i=1}^n p_i \ell(x;\xi_i) - \|\textbf{p}-\frac{\textbf{1}}{n}\|^2 \bigg\}\,,
\end{align}
where $\textbf{p}=(p_1,\cdots,p_n)$, $\mathcal{S} = \{\textbf{p}\in \mathbb{R}^n: \sum_{i=1}^n p_i =1, p_i\geq 0\}$. Here
$\ell(x;\xi_i)$ denotes the loss function over Riemannian manifold $\mathcal{M}$, which applies to many machine learning
problems such as PCA \cite{han2020riemannian},
dictionary learning \cite{sun2016complete}, DNNs \cite{huang2018orthogonal},
structured low-rank matrix learning \cite{jawanpuria2018unified,mao2016novel,vandereycken2013low},
among others.
For example, the task of PCA can be cast on a Grassmann manifold.

Recently some algorithms \cite{ferreira2005singularities,li2009monotone,wang2010monotone} have been studied
for variational inequalities on Riemannian manifolds, which are the implicit minimax problems on Riemannian manifolds.
Meanwhile, some methods \cite{lin2020projection,huang2021riemannian} for computing the
projection robust Wasserstein distance, which can be represented as a minimax optimization over the Stiefel manifold \cite{weed2019sharp}.
To the best of our knowledge, the existing explicitly minimax optimization methods such as gradient descent ascent method only focus on the minimax problems in Euclidean space.

To fill this gap,
in the paper, we study the explicit minimax optimization problems over the general Riemannian manifold, and
propose a class of efficient Riemannian gradient-based algorithms to solve the Geodesically-Nonconvex Strongly-Concave (GNSC) minimax problem (\ref{eq:1})
via using general retraction and vector transport.
When Problem (\ref{eq:1}) is deterministic, we propose a new deterministic Riemannian gradient descent ascent algorithm.
When Problem (\ref{eq:1}) is stochastic (i.e, Problem (\ref{eq:2})), we propose two efficient stochastic Riemannian gradient descent ascent algorithms.
Our main \textbf{contributions} can be summarized as follows:
\begin{itemize}
\setlength{\itemsep}{0pt}
\item[1)] We propose an effective Riemannian gradient descent ascent (RGDA) algorithm for the deterministic minimax Problem (\ref{eq:1}). Moreover, we prove that the RGDA has a sample complexity of $O(\kappa^2\epsilon^{-2})$ in finding an $\epsilon$-stationary solution of Problem (\ref{eq:1}).
\item[2)] Meanwhile, we present an effective Riemannian stochastic gradient descent ascent (RSGDA) algorithm for the stochastic minimax Problem (\ref{eq:2}), which has a sample complexity of $O(\kappa^4\epsilon^{-4})$ in searching for an $\epsilon$-stationary solution of Problem (\ref{eq:2}).
\item[3)] We further propose an accelerated Riemannian stochastic gradient descent ascent (Acc-RSGDA) algorithm based on the variance-reduced technique of STORM \cite{cutkosky2019momentum}. We prove our Acc-RSGDA achieves a lower sample complexity of $\tilde{O}(\kappa^4\epsilon^{-3})$.
\item[4)] Extensive experimental results on the robust DNNs training and distributionally robust optimization over Stiefel manifold demonstrate the efficiency of our proposed algorithms.
\end{itemize}

\begin{figure*}[t]%
    \centering
    \subfloat[Retraction $R_x$]{{\includegraphics[width=0.4\textwidth]{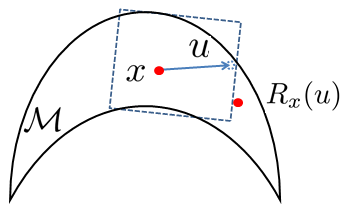} }}%
    \hspace{.6in}
    \subfloat[Vector Transport $\mathcal{T}^y_x$]{{\includegraphics[width=0.4\textwidth]{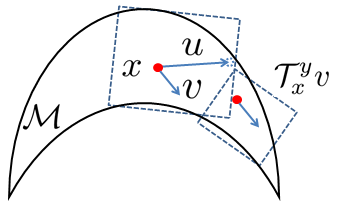} }}%
    \caption{Illustration of manifold operations.(a) A vector $u$ in $T_x\mathcal{M}$ is mapped to $R_x(u)$ in $\mathcal{M}$;
    (b) A vector $v$ in $T_x\mathcal{M}$ is transported to $T_y\mathcal{M}$ by $\mathcal{T}^y_xv$ (or $\mathcal{T}_uv$), where $y=R_x(u)$ and $u\in T_x\mathcal{M}$.}
    \label{fig:1}
\end{figure*}

\section{Related Works}
In this section, we briefly review the minimax optimization and Riemannian manifold optimization, respectively.
\subsection{ Minimax Optimization }
Minimax optimization \cite{razaviyayn2020nonconvex} recently has been widely applied in many machine learning problems such as adversarial training \cite{goodfellow2014generative},
reinforcement learning \cite{zhang2020model},
and robust federated learning \cite{deng2020distributionally}. Meanwhile,
many efficient minimax methods \cite{lin2019gradient,nouiehed2019solving,thekumparampil2019efficient,lin2020near,lu2020hybrid,yang2020global,yang2020catalyst,zhang2021complexity,
yan2020sharp,luo2020stochastic,chen2021proximal,huang2021efficient,huang2022accelerated}
have been proposed for solving these minimax optimization problems.
For example, \cite{thekumparampil2019efficient} proposed a class of efficient dual implicit accelerated gradient
algorithms to solve smooth minimax optimization.
\cite{lin2019gradient} studied the convergence properties of the gradient decent ascent (GDA) methods for
nonconvex minimax optimization. 
Subsequently, the accelerated GDA algorithms \cite{lin2020near} have been proposed for minimax optimization. 
Meanwhile, \cite{yang2020catalyst} presented a catalyst accelerated framework for minimax optimization.  
Moreover, \cite{luo2020stochastic,huang2022accelerated} proposed some faster stochastic variance-reduced GDA algorithms
to solve the stochastic nonconvex-strongly-concave minimax problems.
\cite{yang2020global} studied the convergence propoerties of GDA methods for solving a class of nonconvex-nonconcave minimax problems. 
More recently, a class of efficient mirror descent ascent algorithms \cite{huang2021efficient} have been proposed for nonconvex nonsmooth minimax optimization. 

\subsection{Riemannian Manifold Optimization }
Riemannian manifold optimization methods have been widely applied in machine learning problems including dictionary learning \cite{sun2016complete},
low-rank matrix completion \cite{mao2016novel,vandereycken2013low},
DNNs \cite{huang2018orthogonal} and natural language processing \cite{sakai2021riemannian}.
Many Riemannian optimization methods have been recently proposed. \emph{E.g.} \cite{zhang2016first,liu2017accelerated}
 proposed some efficient
first-order gradient methods for geodesically convex
functions. Subsequently, \cite{zhang2016riemannian} presented fast stochastic
variance-reduced methods to Riemannian manifold optimization.
More recently, \cite{sato2019riemannian} proposed fast
first-order gradient algorithms for Riemannian manifold optimization
by using general retraction and vector transport.
Subsequently, based on these retraction and vector transport,
some fast Riemannian gradient-based methods \cite{zhang2018r,kasai2018riemannian,han2020riemannian,zhou2021faster,han2021improved}
have been proposed for non-convex optimization.
Riemannian Adam-type algorithms \cite{kasai2019riemannian} have been introduced for matrix manifold optimization.
Subsequently, \cite{sakai2021riemannian} proposed an efficient Riemannian adaptive optimization algorithm to
natural language processing.
Meanwhile, some algorithms \cite{ferreira2005singularities,li2009monotone,wang2010monotone} have been studied
for variational inequalities on Riemannian manifolds, which are the implicit minimax problems on Riemannian manifolds. More recently, \cite{huang2022riemannian} studied the stochastic composition optimization on Riemannian manifolds.  

\noindent\textbf{Notations:}
$I_d$ denotes the identity matrix with $d$ dimension. $\mbox{diag}(a) \in \mathbb{R}^{d\times d}$ denotes a diagonal matrix, whose diagonal elements come from vector $a\in \mathbb{R}^d$.  $\mbox{sign}(\cdot)$ denotes the sign function, i.e., if $x>0$, $\mbox{sign}(x)=1$; if $x=0$, $\mbox{sign}(x)=0$; otherwise $\mbox{sign}(x)=-1$.   $\|\cdot\|$ denotes the $\ell_2$ norm for vectors and Frobenius norm for matrices. 
$\langle x,y\rangle$ denotes the inner product of two vectors $x$ and $y$. 
For function $f(x,y)$, $f(x,\cdot)$ denotes  function \emph{w.r.t.} the second variable with fixing $x$,
and $f(\cdot,y)$ denotes function \emph{w.r.t.} the first variable with fixing $y$.
Given a convex closed set $\mathcal{Y}$, we define a projection operation on the set $\mathcal{Y}$ as
$\mathcal{P}_{\mathcal{Y}}(y_0) = \arg\min_{y\in \mathcal{Y}}\frac{1}{2}\|y-y_0\|^2$.
We denote $a=O(b)$ if $a\leq Cb$ for some constant $C>0$, and the notation $\tilde{O}(\cdot)$ hides logarithmic terms.
The operation $\bigoplus$ denotes the Whitney sum that
that takes two vector bundles over a fixed space and produces a new vector bundle over the same space. Given function $f(x)$, let $\mbox{grad} f(x)$ denote its Riemannian gradients at Riemannian manifold and $\nabla f(x)$ denote its  gradients at   Euclidean space. 
Given $\mathcal{B}_t=\{\xi_t^i\}_{i=1}^B$ for any $t\geq 1$,
let $\nabla f_{\mathcal{B}_t}(x) = \frac{1}{B}\sum_{i=1}^B \nabla f(x;\xi_t^i)$ and 
$\mbox{grad}  f_{\mathcal{B}_t}(x) = \frac{1}{B}\sum_{i=1}^B \mbox{grad} f(x;\xi_t^i)$.

\section{Preliminaries}
In this section, we first re-visit some basic information on the Riemannian manifold $\mathcal{M}$.
In general, the manifold $\mathcal{M}$ is endowed with a smooth inner
product $\langle \cdot, \cdot \rangle_x : T_x \mathcal{M} \cdot T_x \mathcal{M} \rightarrow \mathbb{R}$
on tangent space $T_x \mathcal{M}$
for every $x\in \mathcal{M}$. The induced norm $\|\cdot\|_x$ of a tangent vector in $T_x \mathcal{M}$
is associated with the Riemannian metric.
We first define a retraction $R_x : T_x \mathcal{M} \rightarrow \mathcal{M}$ mapping tangent space $T_x \mathcal{M}$
onto $\mathcal{M}$ with a local rigidity
condition that preserves the gradients at $x\in \mathcal{M}$ (please see Fig.1 (a)).
The retraction $R_x$ satisfies all of the following: 1) $R_x(0)=x$, where $0\in T_x \mathcal{M}$; 2) $D R_x(0)=id_{T_x \mathcal{M}}$,
where $DR_x$ denotes the derivative of $R_x$, and $id_{T_x \mathcal{M}}$ denotes an identity mapping on $T_x \mathcal{M}$.
In fact, exponential mapping $\mbox{Exp}_x$ is a special case of
retraction $R_x$ that locally approximates the exponential mapping $\mbox{Exp}_x$ to the first order on the manifold.

Next, we define a vector transport $\mathcal{T}:T \mathcal{M} \bigoplus T \mathcal{M}\rightarrow T \mathcal{M}$ (please see Fig.1 (b))
that satisfies all of the following 1) $\mathcal{T}$ has an associated retraction $R$, \emph{i.e.}, for $x\in \mathcal{M}$ and $w$,
$u \in T_x \mathcal{M} $, $\mathcal{T}_u w$ is a tangent
vector at $R_x(w)$; 2) $\mathcal{T}_0v=v$; 3) $\mathcal{T}_u(av+b w)=a\mathcal{T}_u v + b\mathcal{T}_u w$ for all $a,b \in \mathbb{R}$
a $u,v,w\in T \mathcal{M}$. Vector
transport $\mathcal{T}^y_x v$ or equivalently $\mathcal{T}_{u}v$ with $y=R_x(u)$ transports
$v\in T_x \mathcal{M}$ along the retraction curve defined by direction $u$.
Here we focus on the isometric vector transport $\mathcal{T}^y_x$, which
satisfies $\langle u,v \rangle_x = \langle\mathcal{T}^y_x u, \mathcal{T}^y_x v\rangle_y$ for all $u,v\in T_x \mathcal{M}$.
Based on these definitions,
we provide some standard assumptions about Problems (\ref{eq:1}) and (\ref{eq:2}) .

\begin{assumption}
$\mathcal{X} \subseteq \mathcal{M}$ is compact. Each component function $f(x,y)$ is twice continuously differentiable in both $x\in \mathcal{X}$ and $y\in \mathcal{Y}$, and
there exist constants $L_{11}$, $L_{12}$, $L_{21}$ and $L_{22}$, such that for every $x,x_1,x_2\in \mathcal{X}$
and $y,y_1,y_2\in \mathcal{Y}$, we have
\begin{align}
  &\|\mbox{grad}_x f(x_1,y;\xi) - \mathcal{T}^{x_1}_{x_2}\mbox{grad}_x f(x_2,y;\xi)\| \leq L_{11}\|u\|, \nonumber \\
  & \|\mbox{grad}_x f(x,y_1;\xi) - \mbox{grad}_x f(x,y_2;\xi)\| \leq L_{12}\|y_1-y_2\|, \nonumber \\
  & \|\nabla_y f(x_1,y;\xi) - \nabla_y f(x_2,y;\xi)\| \leq L_{21}\|u\|, \nonumber \\
  & \|\nabla_y f(x,y_1;\xi) - \nabla_y f(x,y_2;\xi)\| \leq L_{22}\|y_1-y_2\|, \nonumber
\end{align}
where $u\in T_{x_1} \mathcal{M}$ and $x_2 =R_{x_1}(u)$.
\end{assumption}

Assumption 1 is commonly used in Riemannian optimization \cite{sato2019riemannian,han2020riemannian}, and minimax optimization \cite{lin2019gradient,luo2020stochastic}. Here, the terms $L_{11}$, $L_{12}$ and $L_{21}$ implicitly contain the curvature information as in \cite{sato2019riemannian,han2020riemannian}.
Specifically, Assumption 1 implies the partial Riemannian gradient $\mbox{grad}_x f(\cdot,y;\xi)$ for all $y\in \mathcal{Y}$ is
$L_{11}$-Lipschitz continuous with respect to retraction as in \cite{han2020riemannian} and the partial gradient $\nabla_y f(x,\cdot;\xi)$
for all $x\in \mathcal{X}$ is $L_{22}$-Lipschitz continuous as in \cite{lin2019gradient}.

To further verify the rationality of Assumption 1, we consider the Stiefel manifold $\mathcal{M} = \{X\in \mathbb{R}^{d\times r} | X^TX=I_r\}$. For notational simplicity, let matrix $X$ instead of the variable $x$ in Assumption 1. Let $\nabla_X f(X,y)$ denote the gradient of $f(X,y)$ on variable $X$ in the Euclidean space, and $\mbox{grad}_X f(X,y)$ denote the Riemannian gradient of $f(X,y)$ on variable $X$ in the Stiefel manifold. Following \cite{li2020efficient}, $\mbox{grad}_X f(X,y)$ can be seen as a projection onto the tangent space $T_X \mathcal{M}$ of Riemannian $\mathcal{M}$ at $X$, which can be computed as follows:
\begin{align}
  \mbox{grad}_X f(X,y) & = P_{T_X}\big(\nabla_X f(X,y)\big)  = WX,  \\
  W & = \hat{W} - \hat{W}^T, \nonumber \\
 \hat{W} & = \nabla_X f(X,y)X^T - \frac{1}{2} X(X^T\nabla_X f(X,y)X^T). \nonumber 
\end{align}
Then we have for any $X_1,X_2 \in \mathcal{M}$,
\begin{align} \label{eq:L1}
 & \|\mbox{grad}_X f(X_1,y) - \mathcal{T}_{X_2}^{X_1}\mbox{grad}_X f(X_2,y)\|  \\
 & = \|P_{T_{X_1}}\big(\nabla_X f(X_1,y)\big) - \mathcal{T}_{x_2}^{x_1}P_{T_{X_2}}\big(\nabla_X f(X_2,y)\big)\|, \nonumber \\
 &  \leq \|\nabla_X f(X_1,y) -\nabla_X f(X_2,y)\| \leq L\|X_1-X_2\|, \nonumber
\end{align}
where the last inequality holds by Lipschitz continuous for gradient in the Euclidean space.
Let $d(X_1,X_2)$ denote geodesic distance between $X_1$ and $X_2$ in $\mathcal{M}$, then we have 
$d(X_1,X_2)=\zeta\|X_1-X_2\|$, where $\zeta>0$ denote curvature parameter of manifold $\mathcal{M}$. 
In our Assumption 1, due to $X_2 = R_{X_1}(u)$, we have $\|u\|= d(X_1,X_2)$. According to the above 
(\ref{eq:L1}), we have 
\begin{align}
& \|\mbox{grad}_X f(X_1,y) - \mathcal{T}_{x_2}^{x_1}\mbox{grad}_X f(X_2,y)\|  \nonumber \\
& \leq L\|X_1-X_2\|= \frac{L}{\zeta}d(X_1,X_2) =\frac{L}{\zeta}\|u\|,
\end{align}
where $X_2 = R_{X_1}(u)$. 
This similarly holds for the other inequalities in our Assumption 1. 

For the deterministic problem, let $f(x,y)$ instead of $f(x,y;\xi)$ 
in Assumption 1. In fact, these Lipschitz continuity assumptions are widely applicable to deep learning architectures~\cite{li2020efficient}.
 Note that in the following experiments, given the DNNs using ReLU, the derivative of ReLU is Lipschitz continuous almost everywhere with an appropriate Lipschitz constant, except for a small neighbourhood around 0, whose measure tends
to 0. Such cases do not affect either analysis in theory or training in practice. 

Since $f(x,y)$ is strongly concave in $y\in \mathcal{Y}$, there exists a unique solution to
the problem $\max_{y\in \mathcal{Y}}f(x,y)$ for any $x$.
We define the function $\Phi(x)=\max_{y\in \mathcal{Y}}f(x,y)$ and $y^*(x)=\arg\max_{y\in \mathcal{Y}}f(x,y)$.
\begin{assumption}
The function $\Phi(x): \mathcal{M} \rightarrow \mathbb{R}$ is $L$-smooth. There exists a constant $L>0$,
for all $x\in \mathcal{X},z=R_x(u)$ with $u\in T_x\mathcal{M}$, such that
\begin{align}
 \Phi(z) \leq \Phi(x) + \langle\mbox{grad}\Phi(x),u\rangle + \frac{L}{2}\|u\|^2. \nonumber
\end{align}
\end{assumption}
\begin{assumption}
The objective function $f(x,y)$ is $\mu$-strongly concave w.r.t $y$,
i.e., for any $x\in \mathcal{M}$, $y_1,y_2\in \mathcal{Y}$
\begin{align}
 f(x,y_1) \leq f(x,y_2) \!+\! \langle\nabla_y f(x,y_2), y_1\!-\!y_2\rangle \!-\! \frac{\mu}{2}\|y_1\!-\!y_2\|^2. \nonumber
\end{align}
\end{assumption}
\begin{assumption}
The function $\Phi(x)$ is bounded from below in $\mathcal{M}$, \emph{i.e.,} $\Phi^* = \inf_{x\in \mathcal{M}} \Phi(x)$.
\end{assumption}
\begin{assumption}
The variance of stochastic gradient is bounded, \emph{i.e.,} there exists a constant $\sigma_1 >0$ such that for all $x$,
it follows $\mathbb{E}_{\xi}\|\mbox{grad}_x f(x,y;\xi) - \mbox{grad}_x f(x,y)\|^2 \leq \sigma_1^2$; There exists a constant $\sigma_2 >0$
such that for all $y$,
it follows $\mathbb{E}_{\xi}\|\nabla_y f(x,y;\xi) - \nabla_y f(x,y)\|^2 \leq \sigma_2^2$.
We also define $\sigma = \max\{\sigma_1, \sigma_2\}$.
\end{assumption}

Assumption 2 imposes the smooth of function $\Phi(x)$ over Riemannian manifold $\mathcal{M}$,
as in  \cite{sato2019riemannian,han2021improved,han2020riemannian}.
Assumption 3 imposes the strongly concave of $f(x,y)$ on variable $y$, as in \cite{lin2019gradient,luo2020stochastic}.
Assumption 4 guarantees the feasibility of the GNSC minimax problem (\ref{eq:1}), as the nonconex-strongly-concave minimax optimization on Euclidean space used in \cite{lin2019gradient,luo2020stochastic}.
Assumption 5 imposes the bounded variance of stochastic (Riemannian) gradients,
which is commonly used in the stochastic optimization \cite{han2021improved,lin2019gradient,luo2020stochastic}.

\setlength{\textfloatsep}{2pt}
\begin{algorithm}[tb]
\caption{ RGDA and RSGDA Algorithms }
\label{alg:1}
\begin{algorithmic}[1] 
\STATE {\bfseries Input:}  $T$, parameters $\{\gamma, \lambda, \eta_t\}_{t=1}^T$, mini-batch size $B$,
and initial input $x_1\in \mathcal{M}$, $y_1\in\mathcal{Y}$; \\
\FOR{$t = 1, 2, \ldots, T$}
\STATE \textbf{(RGDA)} Compute deterministic gradients
\begin{align}
  v_t =  \mbox{grad}_x f(x_t,y_t), \  w_t =  \nabla_y f(x_t,y_t); \nonumber
\end{align}
\STATE \textbf{(RSGDA)} Draw $B$ i.i.d. samples $\{\xi^i_t\}_{i=1}^B$, then compute stochastic gradients
\begin{align}
  &  v_t = \frac{1}{B} \sum_{i=1}^B \mbox{grad}_x f(x_t,y_t;\xi^i_t), \nonumber \\
  &  w_t = \frac{1}{B} \sum_{i=1}^B \nabla_y f(x_t,y_t;\xi^i_t); \nonumber
\end{align}
\STATE Update: $x_{t+1} = R_{x_t}(- \gamma\eta_t v_t)$;
\STATE Update: $\tilde{y}_{t+1} = \mathcal{P}_{\mathcal{Y}}(y_t + \lambda w_t)$ and $y_{t+1} = y_t + \eta_t(\tilde{y}_{t+1}-y_t)$;
\ENDFOR
\STATE {\bfseries Output:}  $x_{\zeta}$ and $y_{\zeta}$ chosen uniformly random from $\{x_t, y_t\}_{t=1}^{T}$.
\end{algorithmic}
\end{algorithm}

\section{Riemanian Gradient-Based Methods}
In this section, we propose a class of Riemannian gradient-based methods to
solve the deterministic and stochastic GNSC minimax problems (\ref{eq:1}) and (\ref{eq:2}) ,
respectively.

\subsection{ RGDA and RSGDA Algorithms }
In this subsection, we propose an efficient Riemannian gradient descent ascent (RGDA) algorithm
to solve the deterministic minimax Problem (\ref{eq:1}).
At the same time, we propose a standard Riemannian stochastic gradient descent ascent (RSGDA) algorithm
to solve the stochastic minimax Problem (\ref{eq:2}).
Algorithm \ref{alg:1} summarizes the algorithmic framework of our RGDA and RSGDA algorithms.

At the line 3 of Algorithm \ref{alg:1}, we calculate the deterministic Riemannian gradient in variable $x\in \mathcal{M}$, and
calculate the deterministic gradient in variable $y\in \mathcal{Y}$.
At the line 4 of Algorithm \ref{alg:1}, we calculate the stochastic Riemannian gradient for variable $x\in \mathcal{M}$, and
calculate the stochastic gradient for variable $y\in \mathcal{Y}$.

At the line 5 of Algorithm \ref{alg:1}, we use the
Riemannian gradient descent to update variable $x$ based on the retraction operator $R_{x_t} (\cdot)$, which guarantees the variable $x_t$ for all $t\geq 1$
in the manifold $\mathcal{M}$. Here $R_{x_t} (\cdot)$ can be seen as a generalized projection operator, which
 can be competent to the general Riemannian manifolds.
For example, we consider the popular Stiefel manifold $\mathcal{M}=\mbox{St}(r,d) = \{X\in \mathbb{R}^{d\times r}\ : \ X^T X = I_r \}$ that is a nonconvex constraint set in the Euclidean space. 
Given $g_t=-\gamma\eta_tv_t \in T_{x_t}\mathcal{M}$, we can define a standard QR-based retraction:  $R_{x_t}(g_t)= QH$, where the matrices $Q$ and $H$ can be obtained from the 
QR decomposition of matrix $x_t+g_t\in \mathbb{R}^{d\times r}$, i.e., $x_t+g_t=QR$, and $H=\mbox{diag}\big(\big\{\mbox{sign}(R_{i,i})\big\}_{i=1}^r\big)$. 
It is well known that the standard projected gradient methods with convergence guarantee require the convex constraint sets
belonging to Euclidean space \cite{huang2013optimization}, while our Riemannian gradient-based methods with convergence guarantee 
do not need the convex constraint sets (Please see the following convergence analysis). 

At the line 6 of Algorithm \ref{alg:1}, we simultaneously use a projection iteration and a momentum iteration to update the variable
$y$, where we use $0<\eta_t\leq 1$ to ensure the variable $y_t$ for all $t\geq 1$
in convex constraint $\mathcal{Y}$. Note that we use two learning rates $\gamma$ and $\eta_t$ at the line 5, where
$\gamma$ is a constant learning rate and $\eta_t$ is a dynamic or constant learning rate with iteration $t$.
Under this case, we can flexibly choose learning rates in practice, and can easily analyze the convergence properties of
our algorithms, where simultaneously Riemannian gradient descent on the variable $x\in \mathcal{M}$ and
gradient ascent on the variable $y\in \mathcal{Y}$.

\subsection{ Acc-RSGDA algorithm }
In this subsection, we propose an accelerated
stochastic Riemannian gradient descent ascent (Acc-RSGDA) algorithm
to solve the stochastic minimax Problem (\ref{eq:2}), which builds on the momentum-based variance reduction technique
of STORM \cite{cutkosky2019momentum}.
Algorithm \ref{alg:2} describes the algorithmic framework of Acc-RSGDA method.

\begin{algorithm}[tb]
\caption{ Acc-RSGDA Algorithm}
\label{alg:2}
\begin{algorithmic}[1] 
\STATE {\bfseries Input:}  $T$, parameters $\{\gamma, \lambda, b, m,c_1,c_2\}$ and initial input $x_1\in \mathcal{M}$ and $y_1\in\mathcal{Y}$; \\
\STATE Draw $B$ i.i.d. samples $\mathcal{B}_1=\{\xi^i_1\}_{i=1}^B$, then compute $v_1 = \mbox{grad}_x f_{\mathcal{B}_1}(x_1,y_1)$ and $w_1 = \nabla_y f_{\mathcal{B}_1}(x_1,y_1)$;\\
\FOR{$t = 1, 2, \ldots, T$}
\STATE Update: $x_{t+1} = R_{x_t}(-\gamma\eta_t v_t)$ with $\eta_t = \frac{b}{(m+t)^{1/3}}$;
\STATE Update: $\tilde{y}_{t+1} = \mathcal{P}_{\mathcal{Y}}(y_t + \lambda w_t)$ and $y_{t+1} = y_t + \eta_t(\tilde{y}_{t+1}-y_t)$;
\STATE Draw $B$ i.i.d. samples $\mathcal{B}_{t+1}=\{\xi^i_{t+1}\}_{i=1}^B$, then compute
\begin{align}
v_{t+1} & = \mbox{grad}_x f_{\mathcal{B}_{t+1}}(x_{t+1},y_{t+1}) + (1-\alpha_{t+1}) \nonumber \\
 & \quad \cdot \mathcal{T}^{x_{t+1}}_{x_t}\big[v_t - \mbox{grad}_x f_{\mathcal{B}_{t+1}}(x_t,y_t)\big], \label{eq:7} \\
w_{t+1} & = \nabla_y f_{\mathcal{B}_{t+1}}(x_{t+1},y_{t+1}) + (1-\beta_{t+1}) \nonumber \\
 & \quad \cdot \big[w_t - \nabla_y f_{\mathcal{B}_{t+1}}(x_t,y_t)\big], \label{eq:8}
\end{align}
where $\alpha_{t+1} = c_1\eta_t^2$ and $\beta_{t+1} = c_2\eta_t^2$.
\ENDFOR
\STATE {\bfseries Output:}  $x_{\zeta}$ and $y_{\zeta}$ chosen uniformly random from $\{x_t, y_t\}_{t=1}^{T}$.
\end{algorithmic}
\end{algorithm}

At the line 4 of Algorithm \ref{alg:2}, we use two learning rates $\gamma$ and $\eta_t$, where
$\gamma$ is a constant learning rate and $\eta_t = \frac{b}{(m+t)^{1/3}}$ is a decreasing learning rate with iteration $t$.
Similarly, we can flexibly choose learning rates in practice, and can easily analyze the convergence properties of
our algorithms, where simultaneously Riemannian gradient descent on the variable $x\in \mathcal{M}$ and
gradient ascent on the variable $y\in \mathcal{Y}$.

At the line 6 of Algorithm \ref{alg:2}, we use the momentum-based variance-reduced technique of STORM to estimate 
stochastic Riemannian gradient $v_t$ defined in \eqref{eq:7}.
where $\alpha_{t+1}\in (0,1]$. When $\alpha_{t+1}=1$, $v_{t+1}=\mbox{grad}_x f_{\mathcal{B}_{t+1}}(x_{t+1},y_{t+1})$ will
degenerate a vanilla stochastic Riemannian gradient estimator;
When $\alpha_{t+1}=0$,
$v_{t+1}=\mbox{grad}_x f_{\mathcal{B}_{t+1}}(x_{t+1},y_{t+1}) - \mathcal{T}^{x_{t+1}}_{x_t}\big(\mbox{grad}_x f_{\mathcal{B}_{t+1}}(x_t,y_t) - v_t\big)$
will degenerate a stochastic Riemannian gradient estimator based on variance-reduced technique of SPIDER \cite{fang2018spider}.
Since our Acc-RSGDA algorithm uses variance-reduced technique
of STORM to estimate the stochastic gradients,
it does not rely on large mini-batch size to guarantee its convergence (Please see the following convergence analysis).

Riemannian gradient $\mbox{grad}_x f_{\mathcal{B}_{t+1}}(x_{t+1},y_{t+1})$ is over the tangent space
$T_{x_{t+1}}\mathcal{M}$, while the Riemannian gradient estimator $\mbox{grad}_x f_{\mathcal{B}_{t+1}}(x_t,y_t) - v_t$
is over the tangent space
$T_{x_t}\mathcal{M}$. In order to feasibility of $v_{t+1}$,
we use the vector transport $\mathcal{T}^{x_{t+1}}_{x_t}$ to project the Riemannian gradient estimator $\mbox{grad}_x f_{\mathcal{B}_{t+1}}(x_t,y_t) - v_t$
into the tangent space
$T_{x_{t+1}}\mathcal{M}$. Thus, we can add the term
$\mbox{grad}_x f_{\mathcal{B}_{t+1}}(x_{t+1},y_{t+1})$ and the term $(1-\alpha_{t+1}) \mathcal{T}^{x_{t+1}}_{x_t}\big[v_t - \mbox{grad}_x f_{\mathcal{B}_{t+1}}(x_t,y_t)\big]$.

\subsection{ Novelties of Our Algorithms }
Compared with the existing Riemannian gradient algorithms \cite{zhang2018r,kasai2018riemannian,han2020riemannian} and 
 minimax optimization algorithms \cite{lin2019gradient,luo2020stochastic}, our algorithms have the following main differences:
\begin{itemize}
 \item[1)] Compared with the existing Riemannian gradient algorithms, our algorithms simultaneously use 
a constant learning rate $\gamma$ and a dynamic or constant learning rate $\eta_t$ at each iteration. This dynamic/constant learning rate $\eta_t$ is the same tuning parameter of the 
\textbf{momentum iteration} in updating variable $y$ (i.e., $y_{t+1}=y_t+\eta_t(\tilde{y}_{t+1}-y_t)$). In other words, the learning rate in updating the variable $x\in \mathcal{M}$ depends on the tuning parameter of the
 momentum iteration in updating dual variable $y$. 
 \item[2)] Compared with the existing minimax optimization algorithms, our algorithms simultaneously use a projection iteration and a momentum iteration to update the variable
$y$. Meanwhile, our algorithms use the Riemannian gradients and retraction operator to update variable $x\in \mathcal{M}$ instead of 
the standard gradients and projection operator used in the existing minimax algorithms. 
\end{itemize}

\begin{figure*}[!t]
  \centering
  \includegraphics[width=0.85\textwidth]{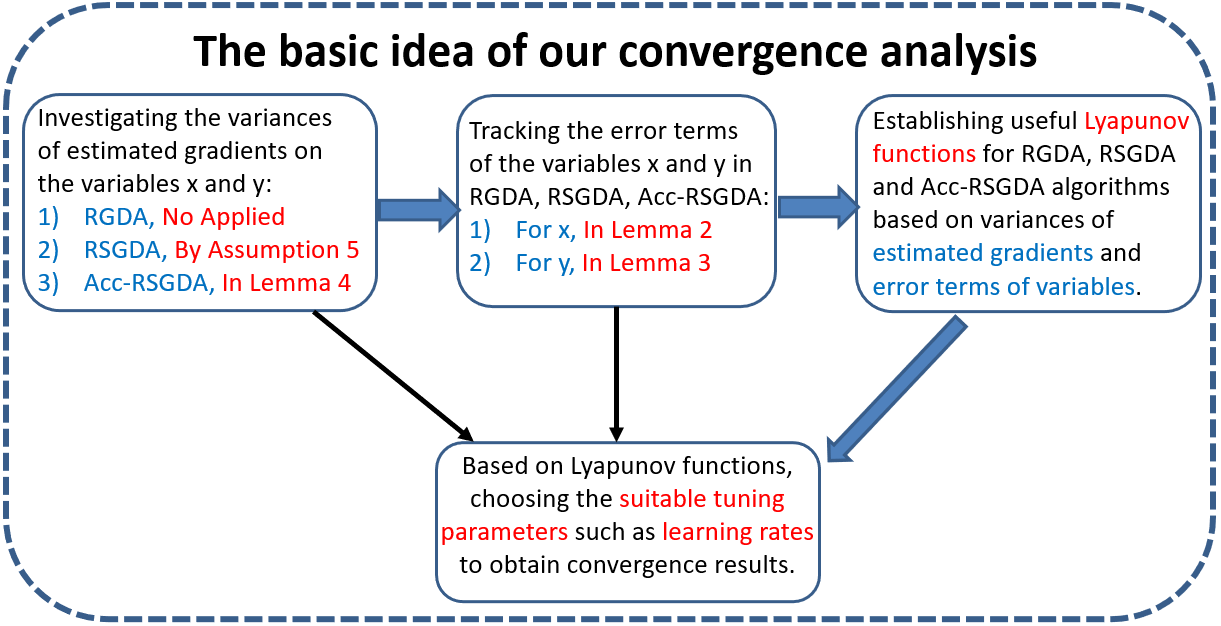}
  \caption{ The basic idea of our convergence analysis. }
  \label{fig:2}
\end{figure*}

\section{Convergence Analysis}
In this section, we study the convergence properties of our RGDA, RSGDA, and Acc-RSGDA algorithms, respectively. 
The basic idea of our convergence analysis is given in Fig. \ref{fig:2}. 
We first give some useful lemmas.

\begin{lemma}
Under the above assumptions, the gradient of function $\Phi(x)=\max_{y\in \mathcal{Y}}f(x,y)$ is
$G$-Lipschitz with respect to retraction, and the mapping or function
$y^*(x)=\arg\max_{y\in \mathcal{Y}}f(x,y)$ is $\kappa$-Lipschitz with respect to retraction.
Given any $x_1, x_2 \in \mathcal{X}\subseteq \mathcal{M}$ and $u\in T_{x_1}\mathcal{M}$, we have:
\begin{align}
& \|\mbox{grad} \Phi(x_1) - \mathcal{T}^{x_1}_{x_2}\mbox{grad} \Phi(x_2)\| \leq G \|u\|,  \\
& \|y^*(x_1)-y^*(x_2)\| \leq \kappa\|u\|,
\end{align}
where $x_2=R_{x_1}(u)$, and $G=\kappa L_{12} + L_{11}$, and $\kappa=L_{21}/\mu$ denotes the number condition of function $f(x,y)$.
\end{lemma}

\begin{lemma}
Suppose the sequence $\{x_t,y_t\}_{t=1}^T$ is generated from Algorithm \ref{alg:1} or \ref{alg:2}.
Given $0<\eta_t\leq \frac{1}{2\gamma L}$,
we have
\begin{align}
  & \Phi(x_{t+1}) \leq \Phi(x_t) + \gamma L_{12}\eta_t \|y^*(x_t)-y_t\|^2 -\frac{\gamma\eta_t}{4}\|v_t\|^2  \nonumber  \\
  & \quad + \gamma\eta_t\|\mbox{grad}_x f(x_t,y_t)-v_t\|^2
  - \frac{\gamma\eta_t}{2}\|\mbox{grad} \Phi(x_t)\|^2.
\end{align}
\end{lemma}

\begin{lemma}
Suppose the sequence $\{x_t,y_t\}_{t=1}^T$ is generated from Algorithm \ref{alg:1} or \ref{alg:2}.
Under the above assumptions, and set $0< \eta_t\leq 1$
and $0<\lambda\leq \frac{1}{6\tilde{L}}$, we have
\begin{align}
     &\|y_{t+1} - y^*(x_{t+1})\|^2 \nonumber \\
     &\leq (1-\frac{\eta_t\mu\lambda}{4})\|y_t -y^*(x_t)\|^2 -\frac{3\eta_t}{4} \|\tilde{y}_{t+1}-y_t\|^2 \nonumber\\
     & + \frac{25\eta_t\lambda}{6\mu}  \|\nabla_y f(x_t,y_t)-w_t\|^2 + \frac{25\gamma^2\kappa^2\eta_t}{6\mu\lambda}\|v_t\|^2,
\end{align}
where $\kappa = L_{21}/\mu$ and $\tilde{L}=\max(1,L_{11},L_{12},L_{21},L_{22})$.
\end{lemma}

Although Problems (\ref{eq:1}) and (\ref{eq:2}) are nonconvex,
following \cite{von2007theory}, there exists a local solution or stationary point $(x^*,y^*)$ satisfies the Nash Equilibrium, i.e.,
$f(x^*,y) \leq f(x^*,y^*) \leq f(x,y^*)$,
where $x^*\in \mathcal{X}\subset \mathcal{M}$ and $y^*\in \mathcal{Y}$.
Here $\mathcal{X}$ is a neighbourhood around an optimal point $x^*$.
Recall that the nonconvex minimax problem \eqref{eq:1} is equivalent to minimizing the nonconvex function $\Phi(x)=\max_{y\in \mathcal{Y}}f(x,y)$ for any $x\in \mathcal{M}$.
It is  NP hard to find the global
minimum of $\Phi(x)$ in general since $\Phi(x)$ is nonconvex in $x\in \mathcal{M}$. Thus, we will find the stationary points of
function $\Phi(x)$, which is equal to the stationary points of the minimax problem \eqref{eq:1}. Next we define an $\epsilon$-stationary point of
 $\Phi(x)$ in $x\in \mathcal{M}$.
\begin{definition}
 A point $x\in \mathcal{M}$ is an $\epsilon$-stationary point ($\epsilon \geq 0$) of
a differentiable function $\Phi(x)$ if $\|\mbox{grad} \ \Phi(x)\| \leq \epsilon$. If $\epsilon=0$, then
$x$ is a stationary point.
\end{definition}

\subsection{ Convergence Analysis of both RGDA and RSGDA Algorithms }
In this subsection, we study the convergence properties of our RGDA and RSGDA algorithms, respectively.

Suppose the sequence $\{x_t,y_t\}_{t=1}^T$ be generated from our RGDA Algorithm, we establish a useful \emph{Lyapunov} function (i.e., potential function) $\Lambda_t$ for convergence analysis of RGDA, 
defined as 
\begin{align}
\Lambda_t = \Phi(x_t) + \frac{6\gamma\tilde{L}^2}{\lambda\mu}\|y_t - y^*(x_t)\|^2, \ \forall t\geq 1.
\end{align}
\begin{theorem} \label{th:1}
Suppose the sequence $\{x_t,y_t\}_{t=1}^T$ is generated from Algorithm \ref{alg:1} by using \textbf{deterministic} gradients.
Given $y_1=y^*(x_1)$, $\eta=\eta_t$ for all $t\geq 1$, $0< \eta \leq \min(1,\frac{1}{2\gamma L})$,
$0< \lambda\leq \frac{1}{6\tilde{L}}$ and $0< \gamma \leq \frac{\mu\lambda}{10\tilde{L}\kappa}$,
we have
\begin{align}
  \frac{1}{T} \sum_{t=1}^T\|\mbox{grad} \ \Phi(x_t)\|
  \leq  \frac{2\sqrt{\Phi(x_1) - \Phi^*}}{\sqrt{\gamma \eta T}}.
\end{align}
\end{theorem}

\begin{remark}
Since $0< \eta \leq \min(1,\frac{1}{2\gamma L})$ and $0< \gamma \leq \frac{\mu\lambda}{10\tilde{L}\kappa}$,
we have $0< \eta\gamma \leq \min(\frac{\mu\lambda}{10\tilde{L}\kappa},\frac{1}{2 L})$. Let $\eta\gamma = \min(\frac{\mu\lambda}{10\tilde{L}\kappa},\frac{1}{2 L})$, 
we have $\eta\gamma=O(\frac{1}{\kappa^2})$.
The RGDA algorithm has convergence rate of $O\big(\frac{\kappa}{T^{1/2}}\big)$.
By $\frac{\kappa}{T^{1/2}} \leq \epsilon$, i.e., $\|\mbox{grad} \ \Phi(x_\zeta)\| \leq \epsilon$, we
choose $T \geq \kappa^2\epsilon^{-2}$. When our RGDA Algorithm solves the deterministic minimax Problem \eqref{eq:1},
we only need one sample to estimate the gradients
$v_t$ and $w_t$ at each iteration, and need $T$ iterations.
Thus, our RGDA reaches a sample complexity of $T=O(\kappa^2\epsilon^{-2})$
for finding an $\epsilon$-stationary point of Problem \eqref{eq:1}. Note that since the function $f(x,y)$ is $\mu$-strongly concave in $y \in \mathcal{Y}$, given any initial input  $x_1$, we can easily obtain $y_1\approx y^*(x_1)$. So we can assume $y_1=y^*(x_1)$. 
\end{remark}

Suppose the sequence $\{x_t,y_t\}_{t=1}^T$ be generated from our RSGDA Algorithm, we establish a useful \emph{Lyapunov} function $\Theta_t$ for convergence analysis of RSGDA, defined as
\begin{align}
\Theta_t = \mathbb{E}\big[\Phi(x_t) + \frac{6\gamma\tilde{L}^2}{\lambda\mu}\|y_t - y^*(x_t)\|^2 \big], \ \forall t\geq 1.
\end{align}
\begin{theorem} \label{th:2}
Suppose the sequence $\{x_t,y_t\}_{t=1}^T$ is generated from Algorithm \ref{alg:1} by using \textbf{stochastic} gradients.
Given $y_1=y^*(x_1)$, $\eta=\eta_t$ for all $t\geq 1$, $0< \eta \leq \min(1,\frac{1}{2\gamma L})$,
$0< \lambda\leq \frac{1}{6\tilde{L}}$ and $0< \gamma \leq \frac{\mu\lambda}{10\tilde{L}\kappa}$,
we have
\begin{align}
  \frac{1}{T} \sum_{t=1}^T\mathbb{E} \|\mbox{grad} \ \Phi(x_t)\|
  & \leq  \frac{2\sqrt{\Phi(x_1) - \Phi^*}}{\sqrt{\gamma \eta T}} + \big(1+ \frac{5\tilde{L}}{\mu}\big)\frac{\sqrt{2}\sigma}{\sqrt{B}}.
\end{align}
\end{theorem}

\begin{remark}
Since $0< \eta \leq \min(1,\frac{1}{2\gamma L})$ and $0< \gamma \leq \frac{\mu\lambda}{10\tilde{L}\kappa}$,
we have $0< \eta\gamma \leq \min(\frac{\mu\lambda}{10\tilde{L}\kappa},\frac{1}{2 L})$. Let $\eta\gamma = \min(\frac{\mu\lambda}{10\tilde{L}\kappa},\frac{1}{2 L})$, we have $\eta\gamma=O(\frac{1}{\kappa^2})$.
Let $B=T$, the RSGDA algorithm has convergence rate of $O\big(\frac{\kappa}{T^{1/2}}\big)$.
By $\frac{\kappa}{T^{1/2}} \leq \epsilon$, i.e., $\mathbb{E}\|\mbox{grad} \ \Phi(x_\zeta)\| \leq \epsilon$, we
choose $T \geq \kappa^2\epsilon^{-2}$. When our RGDA Algorithm solves the stochastic minimax Problem \eqref{eq:2},
we need $B$ samples to estimate the gradients
$v_t$ and $w_t$ at each iteration, and need $T$ iterations.
Thus, the RSGDA  reaches a sample complexity of $BT=O(\kappa^4\epsilon^{-4})$
for finding an $\epsilon$-stationary point of Problem \eqref{eq:2}.
\end{remark}

\subsection{Convergence Analysis of Acc-RSGDA Algorithm}
In the subsection, we provide the convergence properties of our Acc-RSGDA algorithm.

\begin{lemma}
 Suppose the stochastic gradients $v_t$ and $w_t$ is generated from Algorithm \ref{alg:2}, given
 $0<\alpha_{t+1}\leq 1$ and $0<\beta_{t+1}\leq 1$, we have
\begin{align} 
&\mathbb{E} \|\mbox{grad}_x f(x_{t+1},y_{t+1}) - v_{t+1}\|^2\leq 4(1-\alpha_{t+1})^2L^2_{11}\gamma^2\eta^2_t\mathbb{E}\|v_t\|^2 \nonumber \\
& \quad +(1-\alpha_{t+1})^2 \mathbb{E} \|\mbox{grad}_x f(x_t,y_t) -v_t\|^2   \nonumber \\
& \quad + 4(1-\alpha_{t+1})^2L^2_{12}\eta^2_t\mathbb{E}\|\tilde{y}_{t+1}-y_t\|^2 + \frac{2\alpha_{t+1}^2\sigma^2}{B},
\end{align}
\begin{align} 
 &\mathbb{E} \|\nabla_y f(x_{t+1},y_{t+1}) - w_{t+1}\|^2 \leq 4(1-\beta_{t+1})^2L^2_{21}\gamma^2\eta^2_t\mathbb{E}\|v_t\|^2 \nonumber \\
 & \quad + (1-\beta_{t+1})^2 \mathbb{E} \|\nabla_y f(x_t,y_t) -w_t\|^2 \nonumber \\
 & \quad + 4(1-\beta_{t+1})^2L^2_{22}\eta^2_t\mathbb{E}\|\tilde{y}_{t+1}-y_t\|^2 + \frac{2\beta_{t+1}^2\sigma^2}{B}.
\end{align}
\end{lemma}

Assume the sequence $\{x_t,y_t\}_{t=1}^T$ be generated from our Acc-RSGDA Algorithm, we establish a useful \emph{Lyapunov} function  $\Omega_t $ for convergence analysis of 
Acc-RSGDA, defined as
\begin{align}
& \Omega_t  =  \mathbb{E}\big[\Phi(x_t) + \frac{\gamma}{2\lambda\mu\eta_{t-1}}\big( \|\mbox{grad}_x f(x_t,y_t)-v_t\|^2   \\
&\quad + \|\nabla_y f(x_t,y_t)-w_t\|^2 \big) + \frac{6\gamma\tilde{L}^2}{\lambda\mu} \|y_t-y^*(x_t)\|^2\big] , \ \forall t\geq 1.\nonumber
\end{align}

\begin{theorem} \label{th:3}
Suppose the sequence $\{x_t,y_t\}_{t=1}^T$ is generated from Algorithm \ref{alg:2}. Given $y_1=y^*(x_1)$, $c_1 \geq \frac{2}{3b^3} + 2\lambda\mu$,
$c_2 \geq \frac{2}{3b^3} + \frac{50\lambda\tilde{L}^2}{\mu}$, $b>0$, $m\geq \max\big( 2, (\tilde{c}b)^3\big)$,
$0<\gamma \leq \frac{\mu\lambda}{2\kappa\tilde{L}\sqrt{25+4\mu\lambda}}$ and $0<\lambda\leq \frac{1}{6\tilde{L}}$, we have
\begin{align}
  \frac{1}{T} \sum_{t=1}^T \mathbb{E}\|\mbox{grad} \ \Phi(x_t)\|
  \leq \frac{\sqrt{2M'}m^{1/6}}{T^{1/2}} + \frac{\sqrt{2M'}}{T^{1/3}},
\end{align}
where $\tilde{c}=\max(1,c_1,c_2, 2\gamma L)$ and $M' = \frac{2(\Phi(x_1) - \Phi^*)}{\gamma b} + \frac{2\sigma^2}{B\lambda\mu\eta_0b}
 + \frac{2(c_1^2+c_2^2)\sigma^2 b^2}{B\lambda\mu}\ln(m+T)$.
\end{theorem}

\begin{remark}
Let $c_1=\frac{2}{3b^3} + 2\lambda\mu$, $c_2=\frac{2}{3b^3} + \frac{50\lambda\tilde{L}^2}{\mu}$, $\lambda=\frac{1}{6\tilde{L}}$,
$\gamma=\frac{\mu\lambda}{2\kappa\tilde{L}\sqrt{25+4\mu\lambda}}$ and $\eta_0=\frac{b}{m^{1/3}}$.
It is easily verified that $\gamma=O(\frac{1}{\kappa^2})$, $\lambda=O(1)$, $\lambda\mu=O(\frac{1}{\kappa})$,
$c_1=O(1)$, $c_2=O(\kappa)$, $m=O(\kappa^3)$
and $\eta_0=O(\frac{1}{\kappa})$. Without loss of generality, let $T\geq m=O(\kappa^3)$,
we have $M'=O\big(\kappa^2 + \frac{\kappa^2}{B} + \frac{\kappa^3}{B}\ln(T)\big)$.
When $B=\kappa$, we have $M'=O\big(\kappa^2\ln(T)\big)$. Thus,
the Acc-RSGDA algorithm has a convergence rate of $\tilde{O}\big(\frac{\kappa}{T^{1/3}}\big)$.
By $\frac{\kappa}{T^{1/3}} \leq \epsilon$, i.e., $\mathbb{E}\|\mbox{grad} \ \Phi(x_\zeta)\| \leq \epsilon$, we
choose $T \geq  \kappa^3\epsilon^{-3}$. In Algorithm \ref{alg:2}, we require $B$ samples to estimate
the stochastic gradients $v_t$ and $w_t$ at each iteration, and need $T$ iterations.
Thus, the Acc-RSGDA  has a sample complexity of $BT=\tilde{O}\big(\kappa^4\epsilon^{-3}\big)$
for finding an $\epsilon$-stationary point of Problem (\ref{eq:2}). 
 \textbf{Since our Acc-RSGDA algorithm uses variance-reduced technique
of STORM to estimate the stochastic gradients,
it does not rely on large mini-batch size to guarantee its convergence.}
When $B=1$, our Acc-RSGDA algorithm has a convergence rate of $\tilde{O}\big(\frac{\kappa^{3/2}}{T^{1/3}}\big)$,
and has a sample complexity of $BT=\tilde{O}\big(\kappa^{4.5}\epsilon^{-3}\big)$
for finding an $\epsilon$-stationary point.
\end{remark}

\begin{remark}
 In the above theoretical analysis, we only assume the convexity of constraint set $\mathcal{Y}$, while \cite{lin2019gradient}
 not only assume the convexity of set $\mathcal{Y}$, but also assume and use \textbf{its bounded} (i.e., $|\mathcal{Y}|\leq D$, where $D$ is a positive constant.) 
 to guarantee  convergence of the GDA and SGDA algorithms in \cite{lin2019gradient} (Please see Assumption 4.2 in \cite{lin2019gradient}). 
 Clearly, our assumption is milder than \cite{lin2019gradient}.
 When there does not exist a constraint set on parameter $y$, i.e.,$\mathcal{Y}=\mathbb{R}^d$, our RGDA and RSGDA algorithms and theoretical results still work, 
 while \cite{lin2019gradient} can not work.
\end{remark}

\section{Experiments}
In this section, we conduct experiments on two tasks: 1) robust DNNs training over Riemannian manifold and distributionally robust optimization over Riemannian manifold.
In the experiment, we use the  SGDA~\cite{lin2019gradient} and Acc-MDA~\cite{huang2022accelerated} as the comparison baselines. Since the SGDA and Acc-MDA methods are not designed for optimization on Riemanian manifolds, we add the retraction operation (projection-like) at the end of parameter updates.

\begin{table}[t]
  \centering
  \resizebox{0.49\textwidth}{!}{
  \begin{tabular}{c|c|c|c}
  \hline
  datasets & \#samples & \#dimension & \#classes \\ \hline
  \emph{MNIST} & 60,000 &  $28\!\times \!28$ & 10 \\
  \emph{FashionMNIST} & 60,000 &  $28\!\times\!28$ & 10 \\
  \emph{STL-10} (resized) & 5,000 &  $32\!\times\!32\times\!3$ & 10 \\
  \emph{CIFAR-10} & 50,000 & $32\!\times\!32\times\!3$ & 10 \\

  \hline
  \end{tabular}
  }
\caption{Benchmark datasets used in our experiments}
 \label{tab:datasets}
\end{table}

\begin{table}[t]
\centering
\resizebox{0.32\textwidth}{!}{
  \begin{tabular}{c}
    \toprule
        Inputs ($d$ channels) \\
    \midrule
        Conv $d \to 32$, Batchnorm , ReLU\\
        Conv $32 \to 64$, Batchnorm , ReLU\\
        Conv $64 \to 64$, Batchnorm , ReLU\\
    \midrule
        Max Pool\\
    \midrule
        Linear $200\to 200$, ReLU\\
        Linear $200\to C$\\
    \midrule
    Outputs\\
    \bottomrule
  \end{tabular}
}
\caption{The DNN used in our experiments. $C$ is the number of classes, and $d$ is the number of channels for inputs.}
\label{tab:arch}
\end{table}

\begin{table*}[]
    \centering
    \caption{Test accuracy against nature images and different attacks for 
\textbf{MNIST}. All comparison methods are test against PGD$^{40}$~\cite{kurakin2016adversarial}: PGD attack of 40 steps, and FGSM~\cite{goodfellow2014explaining} attacks.}
    \label{tab:attack_MNIST}
    \resizebox{0.89\textwidth}{!}{
    \begin{tabular}{c|c|c|c|c|c|c|c|c|c}
    \hline
        \multirow{2}{*}{Methods} & \multirow{2}{*}{Nat. Img.}  & \multicolumn{4}{c|}{PGD$^{40}$ $L_\infty$} & \multicolumn{4}{c}{FGSM $L_\infty$}\\\cline{3-10}
        &  &$\varepsilon$=0.1&$\varepsilon$=0.2&$\varepsilon$=0.3&$\varepsilon$=0.4&$\varepsilon$=0.1&$\varepsilon$=0.2&$\varepsilon$=0.3&$\varepsilon$=0.4\\
        \hline
        SGDA& 98.94\%
        &85.95\%&82.10\%& 75.64\% &61.95\% &91.20\%&89.06\%&85.67\%& 78.01\% \\
        \hline
        Acc-MDA&99.23\% &
        86.12\% &82.15\%&75.22\%&58.06\%
        &92.25\% &90.29\%&87.11\%&79.56\% \\
        \hline
        RSGDA& 99.22\% &
        87.47\%&84.17\%&78.61\% &64.92\% &93.05\%&91.26\%&87.47\%& 80.51\% \\
        \hline
        Acc-RSGDA& 99.37\%
        & \textbf{90.08\%} &\textbf{87.29\%}&\textbf{82.65\%}& \textbf{73.66\%} &\textbf{93.38\%}&\textbf{91.67\%}&\textbf{88.83\%}& \textbf{82.81\%} \\
        \hline
    \end{tabular}
    }
\end{table*}

\begin{table*}[]
    \centering
    \caption{Test accuracy against nature images and different attacks for \textbf{FashionMNIST}. All comparison methods are test against PGD$^{40}$ and FGSM attacks.}
    \label{tab:attack_FMNIST}
    \resizebox{0.89\textwidth}{!}{
    \begin{tabular}{c|c|c|c|c|c|c|c|c|c}
    \hline
        \multirow{2}{*}{Methods} & \multirow{2}{*}{Nat. Img.}  & \multicolumn{4}{c|}{PGD$^{40}$ $L_\infty$} & \multicolumn{4}{c}{FGSM $L_\infty$}\\\cline{3-10}
        &  &$\varepsilon$=0.05&$\varepsilon$=0.1&$\varepsilon$=0.15&$\varepsilon$=0.2&$\varepsilon$=0.05&$\varepsilon$=0.1&$\varepsilon$=0.15&$\varepsilon$=0.2\\
        \hline
        SGDA& 82.30\% &69.12\%&66.92\%&64.83\% &62.55\% &73.83\%&72.65\%&71.65\%& 70.64\% \\
        \hline
        Acc-MDA& 83.89\% &68.41\%&65.86\%&63.32\% &60.76\% &73.11\%&71.77\%&70.54\%& 69.05\% \\
        \hline
        RSGDA& 83.23\% &70.23\%&67.84\%&65.57\% &63.48\% &75.85\%&74.95\%&74.33\%& 73.97\% \\
        \hline
        Acc-RSGDA& 84.15\% &\textbf{71.03\%}&\textbf{68.99\%}&\textbf{66.07\%} &\textbf{64.35\%} &\textbf{76.01\%}&\textbf{75.44\%}&\textbf{75.08\%}& \textbf{74.44\%} \\
        \hline
    \end{tabular}
    }
\end{table*}

\begin{table*}[]
    \centering
    \caption{Test accuracy against nature images and different attacks for \textbf{CIFAR10}. All comparison methods are tested against PGD$^{40}$ and FGSM attacks.}
    \resizebox{.95\textwidth}{!}{
    \label{tab:attack_C10}
    \begin{tabular}{c|c|c|c|c|c|c|c|c|c}
    \hline
        \multirow{2}{*}{Methods} & \multirow{2}{*}{Nat. Img.}  & \multicolumn{4}{c|}{PGD$^{40}$ $L_\infty$} & \multicolumn{4}{c}{FGSM $L_\infty$}\\\cline{3-10}
        &  &$\varepsilon$=0.005&$\varepsilon$=0.01&$\varepsilon$=0.015&$\varepsilon$=0.02&$\varepsilon$=0.005&$\varepsilon$=0.01&$\varepsilon$=0.015&$\varepsilon$=0.02\\
        \hline
        SGDA& 64.73\%&{38.52\%}&{33.89\%}&{28.97\%} &{22.50\%}& {41.66\%} &{37.92\%}&{34.04\%}&{28.75\%} \\
        \hline
        Acc-MDA& 69.01\% &42.78\%&38.08\%& 32.84\% & 25.76\% &46.18\%& 42.41\%&38.42\%& 32.87\% \\
        \hline        
        RCG& 64.64\% & 39.73\%& 35.47\%&30.97\% &24.67\%  &42.71\%&39.28\%& 35.83\% &31.11\%\\
        \hline
        Acc-RSGDA& 72.18\% &\textbf{46.36\%}&\textbf{41.70\%}&\textbf{36.46\%} &\textbf{29.16\%} & \textbf{49.41\%}&\textbf{45.62\%}&\textbf{41.55\%}&\textbf{35.94\%} \\
        \hline
        
        \hline
    \end{tabular}
    }
\end{table*}

\begin{table*}[]
    \centering
    \caption{Test accuracy against nature images and different attacks for the first 10 classes of \textbf{CIFAR100}. All comparison methods are tested against PGD$^{40}$ and FGSM attacks.}
    \resizebox{.95\textwidth}{!}{
    \label{tab:attack_C100}
    \begin{tabular}{c|c|c|c|c|c|c|c|c|c}
    \hline
        \multirow{2}{*}{Methods} & \multirow{2}{*}{Nat. Img.}  & \multicolumn{4}{c|}{PGD$^{40}$ $L_\infty$} & \multicolumn{4}{c}{FGSM $L_\infty$}\\\cline{3-10}
        &  &$\varepsilon$=0.005&$\varepsilon$=0.01&$\varepsilon$=0.015&$\varepsilon$=0.02&$\varepsilon$=0.005&$\varepsilon$=0.01&$\varepsilon$=0.015&$\varepsilon$=0.02\\
        \hline
        SGDA& 69.60\%&{ 45.62\%}&{41.47\%}&{37.65\%} &{32.30\%}& {47.48\%} &{43.87\%}&{40.40\%}&{ 36.10\%} \\
        \hline
        Acc-MDA& 70.70\% &47.42\%&43.20\%& 38.40\% & 31.90\% & 49.73\%&  46.10\%&42.15\%& 37.10\% \\
        \hline        
        RCG& 69.30\% &  47.27\%&  43.10\%&38.60\% &31.80\%  &49.98\%&  46.67\%&43.25\% &37.40\%\\
        \hline
        Acc-RSGDA& 70.10\% &\textbf{48.12\%}&\textbf{44.27\%}&\textbf{39.60\%} &\textbf{33.50\%} & \textbf{50.52\%}&\textbf{47.30\%}&\textbf{ 43.60\%}&\textbf{38.70\%} \\
        \hline
        
        \hline
    \end{tabular}
    }
\end{table*}

\begin{table*}[]
    \centering
    \caption{Test accuracy against nature images and different attacks for \textbf{STL10}. All comparison methods are tested against PGD$^{40}$ and FGSM attacks.}
    \resizebox{.95\textwidth}{!}{
    \label{tab:attack_S10}
    \begin{tabular}{c|c|c|c|c|c|c|c|c|c}
    \hline
        \multirow{2}{*}{Methods} & \multirow{2}{*}{Nat. Img.}  & \multicolumn{4}{c|}{PGD$^{40}$ $L_\infty$} & \multicolumn{4}{c}{FGSM $L_\infty$}\\\cline{3-10}
        &  &$\varepsilon$=0.005&$\varepsilon$=0.01&$\varepsilon$=0.015&$\varepsilon$=0.02&$\varepsilon$=0.005&$\varepsilon$=0.01&$\varepsilon$=0.015&$\varepsilon$=0.02\\
        \hline
        SGDA& 51.24\%&{26.28\%}&{ 22.53\%}&{18.81\%}&{ 14.12\%} & {28.32\%} &{25.06\%}&{ 21.88\%} &{ 17.81\%} \\
        \hline
        Acc-MDA& 51.94\% &  28.22\%&24.39\%& 20.66\% & 16.04\% &30.28\%& 27.01\%&23.93\%& 19.88\% \\
        \hline        
        RCG& 51.86\% & 28.62\%& 24.83\%&20.96\% &16.54\% & 30.80\% &27.56\%& 24.34\%& 20.31\%\\
        \hline
        Acc-RSGDA& 52.51\% &\textbf{29.48\%}&\textbf{ 25.64\%}&\textbf{21.76\%} &\textbf{ 16.76\%} & \textbf{ 31.48\%}&\textbf{ 28.20\%}&\textbf{ 24.91\%}&\textbf{20.69\%} \\
        \hline
        
        \hline
    \end{tabular}
    }
\end{table*}

\subsection{Robust DNNs Training}
 In this subsection, we focus on the robust DNNs training over Riemannian manifold defined in Problem~\eqref{eq:4}, which is a nonconvex and nonconcave minimax problem. Following~\cite{nouiehed2019solving}, we cast the original robust training problem into the following nonconvex-(strongly)-concave problem:
\begin{align} \label{eq:13}
 \min_{x \in \mathcal{M}}  \max_{u \in \mathcal{U}} & \frac{1}{n}\sum_{i=1}^n\sum_{j=1}^{C} u_j\ell(h(a_{ij}^K;x),b_i) -  r(u),\\
\mbox{s.t.} & \ \mathcal{U}=\{ u \in \mathbb{R}^C\ |\ u\geq0, \ \|u\|_1 = 1\}, \nonumber
\end{align}
where $a_{ij}^K$ is the permuted sample after $K$ iterations of Projected Gradient Descent (PGD)~\cite{kurakin2016adversarial} attack, and $C$ is the number of classes for the dataset. Here $r(u)$ is a (strongly) convex regularization term, e.g., $r(u)=\alpha \|u-1/C\|^2$ or KL
divergence $r(u)=\alpha \sum_{i=1}^Cu_i\log(u_iC)$, where $\alpha\geq 0$ is a tuning parameter. In the experiment, we use Stiefel manifold $\mathcal{M}=\text{St}(r,d) = \{X\in \mathbb{R}^{d\times r}\ : \ X^T X = I_r \}$ on parameters $x$ of DNNs (convolution layers and linear layers).

For robust training, we choose five datasets for this experiment: MNIST, FashionMNIST, CIFAR10, CIFAR100 and STL10. We use a 5 layer DNN as the target model, whose architecture is given in Tab.~\ref{tab:arch}. For five datasets, we set $\{ \gamma,\lambda,\eta_t \} = \{0.1, 0.01, 0.1 \}$ for RSGDA. For SGDA, we set the learning rates of both maximization and minimization as $0.01$. For Acc-RSGDA, we set $\{ \gamma,\lambda, b, m, c_1, c_2 \} = \{1.0, 0.1, 0.5, 8, 512, 512 \}$, and we apply the same hyper-parameters to Acc-MDA to ensure a fair comparison. We set $K=3$ for five datasets, and $\varepsilon$ for the robust training is set to $0.4$, $0.2$, $0.02$, $0.02$ and $0.02$ for MNIST, FashionMNIST, CIFAR10, CIFAR100 and STL10 separately. We further set the mini-batch size as 512, and the model is trained for 200 epochs.

\begin{figure}[t]%
    \centering
    \subfloat[ MNIST]{{\includegraphics[width=0.5\columnwidth]{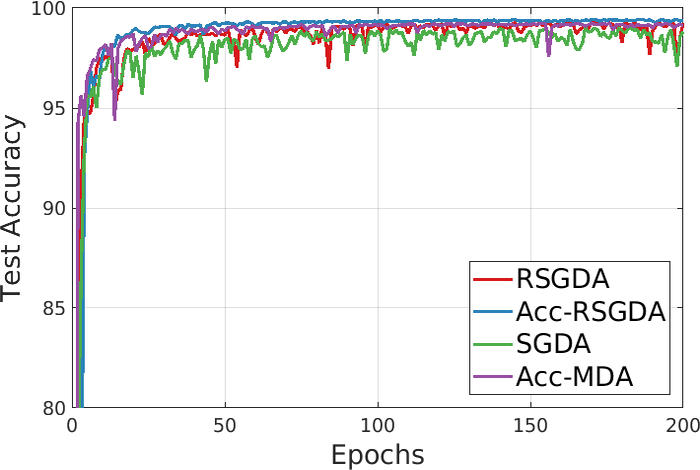} }}%
    \subfloat[ MNIST]{{\includegraphics[width=0.5\columnwidth]{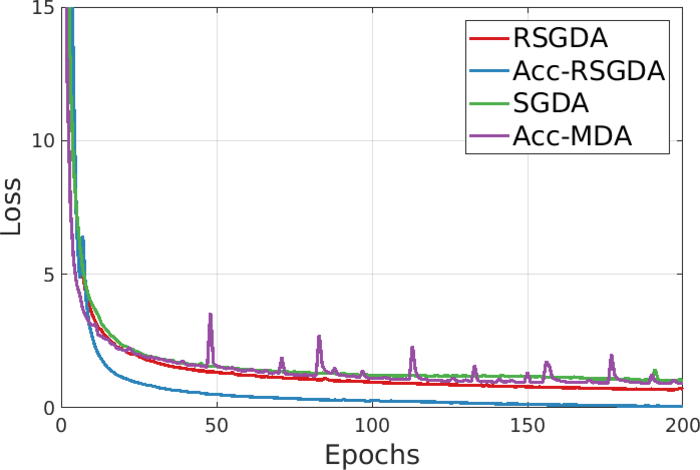} }}%
    \\
    \subfloat[FashionMNIST ]{{\includegraphics[width=0.5\columnwidth]{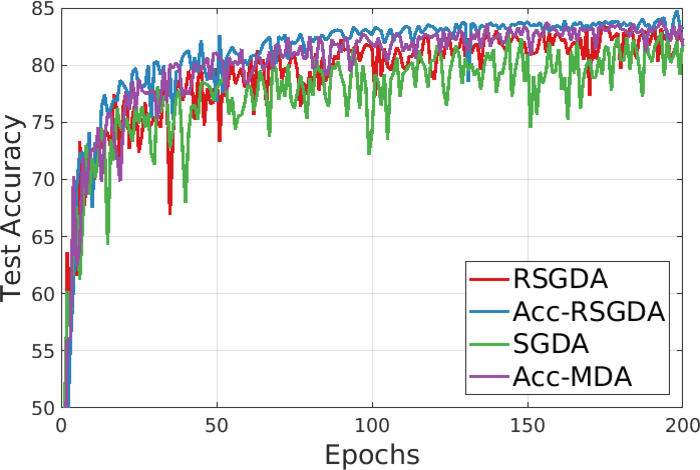} }}%
    \subfloat[FashionMNIST ]{{\includegraphics[width=0.5\columnwidth]{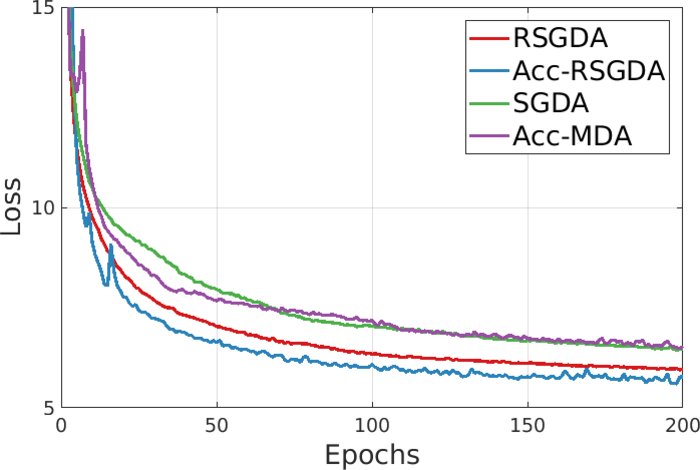} }}%
    \caption{Experimental results for the robust training task. (a, c) Test accuracy with natural images for MNIST and FashionMNIST datasets. (b, d) Training loss for MNIST and FashionMNIST  datasets.}
    \label{fig:robust_train}
\end{figure}

The training progress for robust training is shown in Fig.~\ref{fig:robust_train}. From Fig.~\ref{fig:robust_train}, we can see that our Acc-RSGDA method converges faster than the other comparison baselines, and it can achieve the best test accuracy with natural images for both datasets. RSGDA does not use momentum terms, but it reaches lower training loss compared to SGDA and Acc-MDA. This observation implies that our framework better utilizes the property of Riemannian manifold for robust DNN training. On the other hand, simply adding the retraction operation (Acc-MDA and SGDA) can not achieve the same effect.

The numeric results against different attacks (i.e., PGD attack \cite{kurakin2016adversarial} and Fast Gradient Sign Method (FGSM) attack \cite{goodfellow2014explaining}) are shown in Tab.~\ref{tab:attack_MNIST},  Tab.~\ref{tab:attack_FMNIST}, Tab.~\ref{tab:attack_C10}, Tab.~\ref{tab:attack_C100} and Tab.~\ref{tab:attack_S10}.
Specifically, in the training progress, we report the numeric results against PGD attack of 40 steps and FGSM attack. For all settings, our Acc-RSGDA method achieves the best accuracy against PGD and FGSM attacks. Interestingly, the Acc-MDA method performs worse than SGDA under PGD and FGSM attacks, which suggests that the momentum may be not functional properly without considering the property of Riemannian manifold.

\begin{figure}[t]%
    \centering
    \subfloat[STL-10]{{\includegraphics[width=0.5\columnwidth]{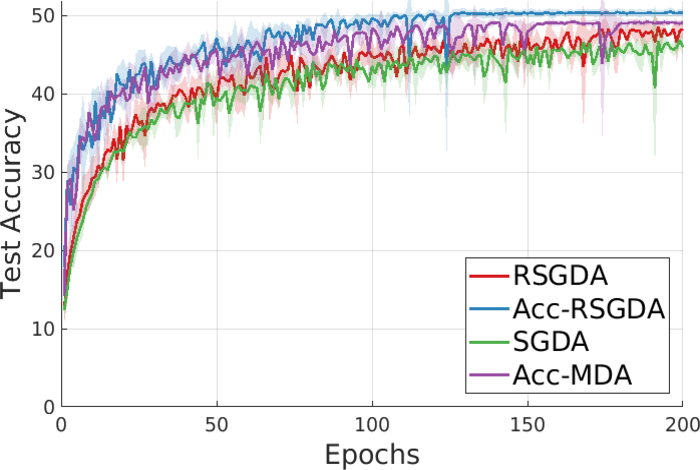} }}%
    \subfloat[STL-10]{{\includegraphics[width=0.5\columnwidth]{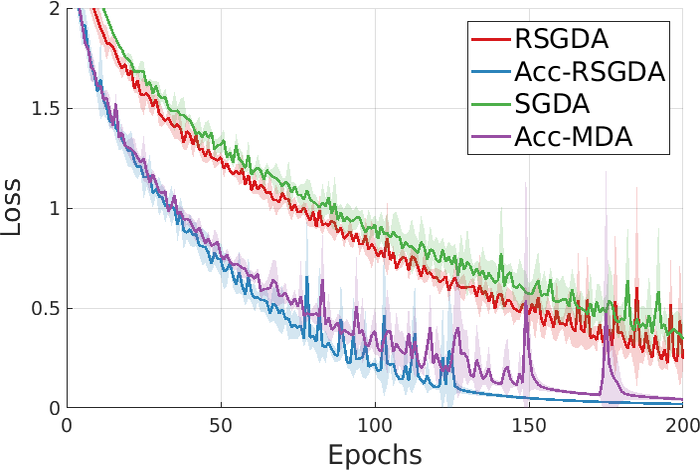} }}%
    \\
    \subfloat[CIFAR-10]{{\includegraphics[width=0.5\columnwidth]{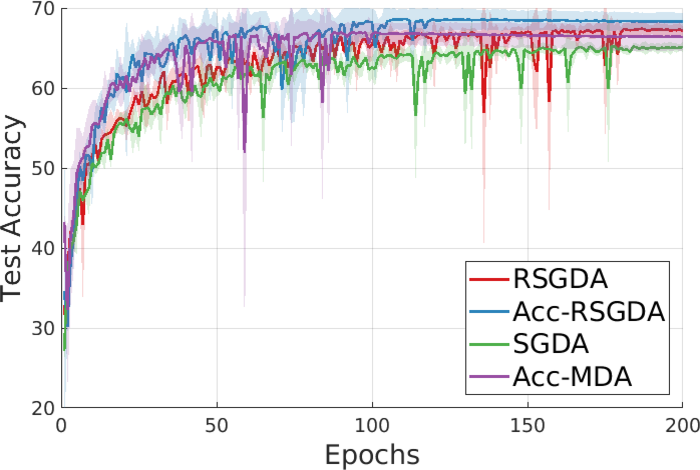} }}%
    \subfloat[CIFAR-10]{{\includegraphics[width=0.5\columnwidth]{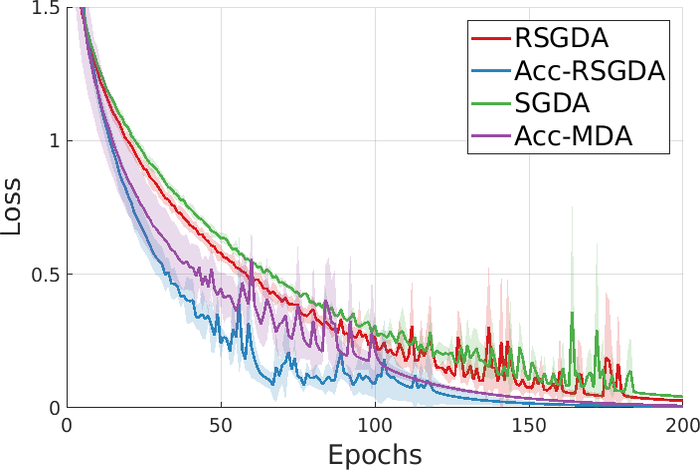} }}%
    \caption{Experimental results for the distributionally robust optimization task. (a, c) Test accuracy for  STL-10 and  CIFAR-10 datasets. (b, d) Training loss for  STL-10 and  CIFAR-10 datasets.}
    \label{fig:dro}
\end{figure}

\subsection{Distributionally Robust Optimization}
In the subsection, we focus on distributionally robust optimization over Riemannian manifold defined in Problem \eqref{eq:6}. CIFAR-10 and STL-10 are selected as the datasets for this task. We use the same DNN architecture from the above robust DNN training for this task. We also apply Stiefel manifold $\mathcal{M}=\text{St}(r,d) = \{X\in \mathbb{R}^{d\times r}\ : \ X^T X = I_r \}$ to the parameters of the DNN. We use the same hyper-parameter setting for RSGDA, Acc-RSGDA, SGDA and Acc-MDA from this task. The mini-batch size is also set 512, and the model is trained for 200 epochs. We report mean and variance across 3 runs for this experiment.

The results are reported in Fig.~\ref{fig:dro}, and shaded areas represent variance. From the figure, we can see that our Acc-RSGDA achieves the best test accuracy and converges fastest.
The difference between Acc-RSGDA and Acc-MDA is small, but due to using the property of Riemannian manifold, Acc-RSGDA is more stable compared to Acc-MDA.

\section{Conclusion}
In the paper, we investigated a class of useful minimax optimization problems on Riemanian manifolds.
Meanwhile, we proposed a class of effective and efficient Riemanian gradient descent ascent algorithms to solve these minimax problems. Moreover, we
studied convergence properties of our proposed algorithms. 
To the best of our knowledge, our Riemannian gradient-based methods are the first to study the minimax optimization
over the \textbf{general} Riemanian manifolds.


%

\ifCLASSOPTIONcompsoc
  \section*{Acknowledgments}
\else
  \section*{Acknowledgment}
\fi

We thank the anonymous reviewers for their helpful comments. We also thank so much for the help of Prof. Heng Huang. 
This work was partially supported by NSFC under Grant No.
61806093. Feihu Huang is the corresponding author.

\ifCLASSOPTIONcaptionsoff
  \newpage
\fi

\bibliographystyle{IEEEtran}
\bibliography{IEEEabrv,RGDA}

\vfill


\begin{onecolumn}
\appendices

\begin{appendices}

\section{Detailed Proofs in Convergence Analysis}
In this section, we provide the detailed convergence analysis of our algorithms.
We first review some useful lemmas.

\begin{lemma} \label{lem:A1}
\cite{nesterov2018lectures} Assume that $f(x)$ is a differentiable convex function and $\mathcal{X}$ is a convex set.
 $x^* \in \mathcal{X}$ is the solution of the
constrained problem $\min_{x\in \mathcal{X}}f(x)$, if
\begin{align}
 \langle \nabla f(x^*), x-x^*\rangle \geq 0, \ \forall x\in \mathcal{X}.
\end{align}
\end{lemma}

\begin{lemma} \label{lem:A2}
\cite{nesterov2018lectures} Assume the function $f(x)$ is $L$-smooth, i.e., $\|\nabla f(x)-\nabla f(y)\|\leq L\|x-y\|$, and
then the following inequality holds
\begin{align}
 |f(y)-f(x)-\nabla f(x)^T(y-x)|\leq \frac{L}{2}\|x-y\|^2.
\end{align}
\end{lemma}

\begin{lemma} \label{lem:A3}
(Restatement of Lemma 1) 
 The gradient of function $\Phi(x)=\max_{y\in \mathcal{Y}}f(x,y)$ is retraction $G$-Lipschitz, and the mapping or function
 $y^*(x)=\arg\max_{y\in \mathcal{Y}}f(x,y)$ is retraction $\kappa$-Lipschitz.
 Given any $x_1, x_2=R_{x_1}(u) \in \mathcal{X}\subset \mathcal{M}$ and $u\in T_{x_1}\mathcal{M}$, we have
 \begin{align}
  & \|\mbox{grad} \Phi(x_1) - \mathcal{T}^{x_1}_{x_2}\mbox{grad} \Phi(x_2)\| \leq G \|u\|, \nonumber \\
  & \|y^*(x_1)-y^*(x_2)\| \leq \kappa\|u\|, \nonumber
 \end{align}
 where $G=\kappa L_{12} + L_{11}$ and $\kappa=L_{21}/\mu$, and vector transport $\mathcal{T}^{x_1}_{x_2}$ transport
 the tangent space of $x_1$ to that of $x_2$.
\end{lemma}

\begin{proof}
 Given any $x_1, x_2=R_{x_1}(u) \in \mathcal{X}$ and $u\in T_{x_1}\mathcal{M}$, define $y^*(x_1)=\arg\max_{y\in \mathcal{Y}} f(x_1,y)$ and $y^*(x_2)=\arg\max_{y\in \mathcal{Y}} f(x_2,y)$,
 by the above Lemma \ref{lem:A1}, we have
 \begin{align}
  & (y-y^*(x_1))^T\nabla_y f(x_1,y^*(x_1)) \leq 0, \quad \forall y\in \mathcal{Y} \label{eq:A1} \\
  & (y-y^*(x_2))^T\nabla_y f(x_2,y^*(x_2)) \leq 0, \quad \forall y\in \mathcal{Y}.  \label{eq:A2}
 \end{align}
 Let $y=y^*(x_2)$ in the inequality (\ref{eq:A1}) and $y=y^*(x_1)$ in the inequality (\ref{eq:A2}),
 then summing these inequalities, we have
 \begin{align} \label{eq:A3}
  (y^*(x_2)-y^*(x_1))^T\big(\nabla_y f(x_1,y^*(x_1)) - \nabla_y f(x_2,y^*(x_2))\big) \leq 0.
 \end{align}
 Since the function $f(x_1,\cdot)$ is $\mu$-strongly concave, we have
 \begin{align}
  & f(x_1,y^*(x_1)) \leq f(x_1,y^*(x_2)) + (\nabla_yf(x_1,y^*(x_2)))^T(y^*(x_1)-y^*(x_2)) - \frac{\mu}{2}\|y^*(x_1)-y^*(x_2)\|^2, \label{eq:A4} \\
  & f(x_1,y^*(x_2)) \leq f(x_1,y^*(x_1)) + (\nabla_yf(x_1,y^*(x_1)))^T(y^*(x_2)-y^*(x_1)) - \frac{\mu}{2}\|y^*(x_1)-y^*(x_2)\|^2. \label{eq:A5}
 \end{align}
 Combining the inequalities (\ref{eq:A4}) with (\ref{eq:A5}), we obtain
 \begin{align} \label{eq:A6}
  (y^*(x_2)-y^*(x_1))^T \big( \nabla_yf(x_1,y^*(x_2))-\nabla_yf(x_1,y^*(x_1))\big) + \mu \|y^*(x_1)-y^*(x_2)\|^2\leq 0.
 \end{align}

 By plugging the inequalities (\ref{eq:A3}) into (\ref{eq:A6}), we have
 \begin{align}
  \mu \|y^*(x_1)-y^*(x_2)\|^2 &\leq (y^*(x_2)-y^*(x_1))^T \big( \nabla_yf(x_2,y^*(x_2))-\nabla_yf(x_1,y^*(x_2))\big) \nonumber \\
  & \leq \|y^*(x_2)-y^*(x_1)\|\|\nabla_yf(x_2,y^*(x_2))-\nabla_yf(x_1,y^*(x_2))\| \nonumber \\
  &\leq L_{21}\|u\|\|y^*(x_2)-y^*(x_1)\|,
 \end{align}
 where the last inequality is due to Assumption 1.
 Thus, we have
 \begin{align}
  \|y^*(x_1)-y^*(x_2)\| \leq \kappa\|u\|,
 \end{align}
 where $\kappa=L_{21}/\mu$ and $x_2=R_{x_1}(u)$, $u\in T_{x_1} \mathcal{M}$.

 Since $\Phi(x)=f(x,y^*(x))$, we have $\mbox{grad} \Phi(x)= \mbox{grad}_x f(x,y^*(x))$. Then we have
 \begin{align}
  & \|\mbox{grad} \Phi(x_1) \!-\! \mathcal{T}^{x_1}_{x_2}\mbox{grad} \Phi(x_2)\| \nonumber \\
  & = \|\mbox{grad}_x f(x_1,y^*(x_1)) - \mathcal{T}^{x_1}_{x_2}\mbox{grad}_x f(x_2,y^*(x_2))\| \nonumber \\
  & \leq \|\mbox{grad}_x f(x_1,y^*(x_1)) \!-\! \mbox{grad}_x f(x_1,y^*(x_2))\| \!+\! \|\mbox{grad}_x f(x_1,y^*(x_2)) \!-\! \mathcal{T}^{x_1}_{x_2}\mbox{grad}_x f(x_2,y^*(x_2))\| \nonumber \\
  & \leq L_{12}\|y^*(x_1)-y^*(x_2)\| + L_{11}\|u\| \nonumber \\
  & \leq (\kappa L_{12} + L_{11}) \|u\|,
 \end{align}
 where $u\in T_{x_1} \mathcal{M}$.

\end{proof}

\begin{lemma} \label{lem:A4}
(Restatement of Lemma 2) 
Suppose the sequence $\{x_t,y_t\}_{t=1}^T$ is generated from Algorithm \ref{alg:1} or \ref{alg:2}.
Given $0<\eta_t\leq \frac{1}{2\gamma L}$,
we have
\begin{align}
  \Phi(x_{t+1}) &\leq \Phi(x_t) + \gamma L_{12}\eta_t \|y^*(x_t)-y_t\|^2 + \gamma\eta_t\|\mbox{grad}_x f(x_t,y_t)-v_t\|^2
  - \frac{\gamma\eta_t}{2}\|\mbox{grad} \Phi(x_t)\|^2 \nonumber \\
  &\quad -\frac{\gamma\eta_t}{4}\|v_t\|^2.
\end{align}
\end{lemma}

\begin{proof}
 According to Assumption 2, i.e., the function $\Phi(x)$ is retraction $L$-smooth,
 we have
 \begin{align}
  \Phi(x_{t+1}) &\leq \Phi(x_t) - \gamma\eta_t\langle\mbox{grad} \Phi(x_t),v_t\rangle + \frac{\gamma^2\eta_t^2L}{2}\|v_t\|^2   \\
  & = \Phi(x_t) + \frac{\gamma\eta_t}{2}\|\mbox{grad} \Phi(x_t)-v_t\|^2 - \frac{\gamma\eta_t}{2}\|\mbox{grad} \Phi(x_t)\|^2
  + (\frac{\gamma^2\eta_t^2 L}{2}-\frac{\gamma\eta_t}{2})\|v_t\|^2 \nonumber \\
  & = \Phi(x_t) + \frac{\gamma\eta_t}{2}\|\mbox{grad} \Phi(x_t)-\mbox{grad}_x f(x_t,y_t) + \mbox{grad}_x f(x_t,y_t)-v_t\|^2
  - \frac{\gamma\eta_t}{2}\|\mbox{grad} \Phi(x_t)\|^2 \nonumber \\
  & \quad + (\frac{\gamma^2\eta_t^2 L}{2}-\frac{\gamma\eta_t}{2})\|v_t\|^2 \nonumber \\
  & \leq \Phi(x_t) + \gamma\eta_t\|\mbox{grad} \Phi(x_t)-\mbox{grad}_x f(x_t,y_t)\|^2 + \gamma\eta_t\|\mbox{grad}_x f(x_t,y_t)-v_t\|^2
  - \frac{\gamma\eta_t}{2}\|\mbox{grad} \Phi(x_t)\|^2 \nonumber \\
  & \quad + (\frac{L\gamma^2\eta_t^2}{2}-\frac{\gamma\eta_t}{2})\|v_t\|^2 \nonumber \\
  & \leq  \Phi(x_t) + \gamma\eta_t\|\mbox{grad} \Phi(x_t)-\mbox{grad}_x f(x_t,y_t)\|^2 + \gamma\eta_t\|\mbox{grad}_x f(x_t,y_t)-v_t\|^2 - \frac{\gamma\eta_t}{2}\|\mbox{grad} \Phi(x_t)\|^2 \nonumber \\
  & \quad -\frac{\gamma\eta_t}{4}\|v_t\|^2, \nonumber
 \end{align}
 where the last inequality is due to $0<\eta_t \leq \frac{1}{2\gamma L}$.

 Considering an upper bound of $\|\mbox{grad} \Phi(x_t)-\mbox{grad}_x f(x_t,y_t)\|^2$, we have
 \begin{align}
  \|\mbox{grad} \Phi(x_t)-\mbox{grad}_x f(x_t,y_t)\|^2 & = \|\mbox{grad}_x f(x_t,y^*(x_t))-\mbox{grad}_x f(x_t,y_t)\|^2 \nonumber \\
  & \leq L_{12} \|y^*(x_t)-y_t\|^2.
 \end{align}
 Then we have
 \begin{align}
  \Phi(x_{t+1}) & \leq \Phi(x_t) + \gamma\eta_tL_{12} \|y^*(x_t)-y_t\|^2 + \gamma\eta_t\|\mbox{grad}_x f(x_t,y_t)-v_t\|^2
  - \frac{\gamma\eta_t}{2}\|\mbox{grad} \Phi(x_t)\|^2 \nonumber \\
  & \quad -\frac{\gamma\eta_t}{4}\|v_t\|^2.
 \end{align}

\end{proof}

\begin{lemma} \label{lem:A5}
(Restatement of Lemma 3) 
Suppose the sequence $\{x_t,y_t\}_{t=1}^T$ is generated from Algorithm \ref{alg:1} or \ref{alg:2}.
Under the above assumptions, and set $0< \eta_t\leq 1$
and $0<\lambda\leq \frac{1}{6\tilde{L}}$, we have
\begin{align}
     \|y_{t+1} - y^*(x_{t+1})\|^2 &\leq (1-\frac{\eta_t\mu\lambda}{4})\|y_t -y^*(x_t)\|^2 -\frac{3\eta_t}{4} \|\tilde{y}_{t+1}-y_t\|^2 \nonumber \\
     & \quad + \frac{25\eta_t\lambda}{6\mu}  \|\nabla_y f(x_t,y_t)-w_t\|^2 +  \frac{25\gamma^2\kappa^2\eta_t}{6\mu\lambda}\|v_t\|^2,
\end{align}
where $\kappa = L_{21}/\mu$ and $\tilde{L}=\max(1,L_{11},L_{12},L_{21},L_{22})$.
\end{lemma}

\begin{proof}
Since the constraint set $\mathcal{Y}$ in Euclidean space,  this proof can easily follow the proofs of Lemma 28 in \cite{huang2022accelerated}.
 According to Assumption 3, i.e., the function $f(x,y)$ is $\mu$-strongly concave w.r.t $y$,
 we have
 \begin{align} \label{eq:E1}
  f(x_t,y) & \leq f(x_t,y_t) + \langle\nabla_y f(x_t,y_t), y-y_t\rangle - \frac{\mu}{2}\|y-y_t\|^2 \nonumber \\
  & = f(x_t,y_t) + \langle w_t, y-\tilde{y}_{t+1}\rangle + \langle\nabla_y f(x_t,y_t)-w_t, y-\tilde{y}_{t+1}\rangle \nonumber \\
  & \quad +\langle\nabla_y f(x_t,y_t), \tilde{y}_{t+1}-y_t\rangle- \frac{\mu}{2}\|y-y_t\|^2.
 \end{align}
 According to the assumption 1, i.e., the function $f(x,y)$ is $L_{22}$-smooth w.r.t $y$, and
 $\tilde{L}\geq L_{22}$, we have
 \begin{align} \label{eq:E2}
  f(x_t,\tilde{y}_{t+1}) - f(x_t,y_{t}) -\langle\nabla_y f(x_t,y_t), \tilde{y}_{t+1}-y_t\rangle
  & \geq -\frac{L_{22}}{2}\|\tilde{y}_{t+1}-y_t\|^2 \nonumber \\
  & \geq -\frac{\tilde{L}}{2}\|\tilde{y}_{t+1}-y_t\|^2 .
 \end{align}
 Combining the inequalities (\ref{eq:E1}) with (\ref{eq:E2}), we have
 \begin{align} \label{eq:E3}
  f(x_t,y) & \leq f(x_t,\tilde{y}_{t+1}) + \langle w_t, y-\tilde{y}_{t+1}\rangle + \langle\nabla_y f(x_t,y_t)-w_t,
  y-\tilde{y}_{t+1}\rangle \nonumber \\
  & \quad - \frac{\mu}{2}\|y-y_t\|^2 + \frac{\tilde{L}}{2}\|\tilde{y}_{t+1}-y_t\|^2.
 \end{align}

 According to the line 6 of Algorithm \ref{alg:1} or \ref{alg:2}, we have $\tilde{y}_{t+1}=\mathcal{P}_{\mathcal{Y}}(y_t + \lambda w_t)
  =\arg\min_{y\in\mathcal{Y}}\frac{1}{2}\|y- y_t - \lambda w_t\|^2$. Since $\mathcal{Y}$ is a convex set
  and the function $\frac{1}{2}\|y - y_t - \lambda w_t\|^2$ is convex, according to Lemma \ref{lem:A1},
  we have
  \begin{align}
  \langle \tilde{y}_{t+1}- y_t - \lambda w_t, y-\tilde{y}_{t+1}\rangle \geq 0, \ y\in \mathcal{Y}.
  \end{align}
 Then we obtain
 \begin{align} \label{eq:E4}
  \langle w_t, y-\tilde{y}_{t+1}\rangle & \leq \frac{1}{\lambda}\langle \tilde{y}_{t+1}- y_t, y-\tilde{y}_{t+1}\rangle \nonumber \\
  & = \frac{1}{\lambda}\langle \tilde{y}_{t+1}- y_t, y_t-\tilde{y}_{t+1}\rangle + \frac{1}{\lambda}\langle \tilde{y}_{t+1}- y_t, y-y_t\rangle \nonumber \\
  & = -\frac{1}{\lambda}\|\tilde{y}_{t+1}- y_t\|^2 + \frac{1}{\lambda}\langle \tilde{y}_{t+1}- y_t, y-y_t\rangle.
  \end{align}
 Combining the inequalities (\ref{eq:E3}) with (\ref{eq:E4}), we have
 \begin{align}
  f(x_t,y) & \leq f(x_t,\tilde{y}_{t+1}) + \frac{1}{\lambda}\langle \tilde{y}_{t+1}- y_t, y-y_t\rangle + \langle\nabla_y f(x_t,y_t)-w_t, y-\tilde{y}_{t+1}\rangle \nonumber \\
  & \quad -\frac{1}{\lambda}\|\tilde{y}_{t+1}- y_t\|^2- \frac{\mu}{2}\|y-y_t\|^2 + \frac{\tilde{L}}{2}\|\tilde{y}_{t+1}-y_t\|^2.
 \end{align}
 Let $y=y^*(x_t)$ and we obtain
 \begin{align}
  f(x_t,y^*(x_t)) & \leq f(x_t,\tilde{y}_{t+1}) + \frac{1}{\lambda}\langle \tilde{y}_{t+1}- y_t, y^*(x_t)-y_t\rangle
  + \langle\nabla_y f(x_t,y_t)-w_t, y^*(x_t)-\tilde{y}_{t+1}\rangle \nonumber \\
  & \quad -\frac{1}{\lambda}\|\tilde{y}_{t+1}- y_t\|^2- \frac{\mu}{2}\|y^*(x_t)-y_t\|^2 + \frac{\tilde{L}}{2}\|\tilde{y}_{t+1}-y_t\|^2.
 \end{align}
 Due to the concavity of $f(\cdot,y)$ and $y^*(x_t) =\arg\max_{y\in \mathcal{Y}} f(x_t,y)$, we have $f(x_t,y^*(x_t)) \geq f(x_t,\tilde{y}_{t+1})$.
 Thus, we obtain
 \begin{align} \label{eq:E5}
  0 & \leq  \frac{1}{\lambda}\langle \tilde{y}_{t+1}- y_t, y^*(x_t)-y_t\rangle
  + \langle\nabla_y f(x_t,y_t)-w_t, y^*(x_t)-\tilde{y}_{t+1}\rangle \nonumber \\
  & \quad -(\frac{1}{\lambda}-\frac{\tilde{L}}{2})\|\tilde{y}_{t+1}- y_t\|^2- \frac{\mu}{2}\|y^*(x_t)-y_t\|^2.
 \end{align}

 By $y_{t+1} = y_t + \eta_t(\tilde{y}_{t+1}-y_t) $, we have
 \begin{align}
  \|y_{t+1}-y^*(x_t)\|^2 & = \|y_t + \eta_t(\tilde{y}_{t+1}-y_t) -y^*(x_t)\|^2 \nonumber \\
  & = \|y_t -y^*(x_t)\|^2 + 2\eta_t\langle \tilde{y}_{t+1}-y_t, y_t -y^*(x_t)\rangle + \eta_t^2\|\tilde{y}_{t+1}-y_t\|^2.
 \end{align}
 Then we obtain
 \begin{align} \label{eq:E6}
  \langle \tilde{y}_{t+1}-y_t, y^*(x_t) - y_t\rangle \leq \frac{1}{2\eta_t}\|y_t -y^*(x_t)\|^2 + \frac{\eta_t}{2}\|\tilde{y}_{t+1}-y_t\|^2 - \frac{1}{2\eta_t}\|y_{t+1}-y^*(x_t)\|^2.
 \end{align}
 Consider the upper bound of the term $\langle\nabla_y f(x_t,y_t)-w_t, y^*(x_t)-\tilde{y}_{t+1}\rangle$, we have
 \begin{align} \label{eq:E7}
  &\langle\nabla_y f(x_t,y_t)-w_t, y^*(x_t)-\tilde{y}_{t+1}\rangle \nonumber \\
  & = \langle\nabla_y f(x_t,y_t)-w_t, y^*(x_t)-y_t\rangle + \langle\nabla_y f(x_t,y_t)-w_t, y_t-\tilde{y}_{t+1}\rangle \nonumber \\
  & \leq \frac{1}{\mu} \|\nabla_y f(x_t,y_t)-w_t\|^2 + \frac{\mu}{4}\|y^*(x_t)-y_t\|^2 + \frac{1}{\mu} \|\nabla_y f(x_t,y_t)-w_t\|^2 + \frac{\mu}{4}\|y_t-\tilde{y}_{t+1}\|^2 \nonumber \\
  & = \frac{2}{\mu} \|\nabla_y f(x_t,y_t)-w_t\|^2 + \frac{\mu}{4}\|y^*(x_t)-y_t\|^2 + \frac{\mu}{4}\|y_t-\tilde{y}_{t+1}\|^2.
 \end{align}

 By plugging the inequalities (\ref{eq:E5}), (\ref{eq:E6}) to (\ref{eq:E7}),
 we have
 \begin{align}
  \frac{1}{2\eta_t\lambda}\|y_{t+1}-y^*(x_t)\|^2 & \leq ( \frac{1}{2\eta_t\lambda}-\frac{\mu}{4})\|y_t -y^*(x_t)\|^2 + ( \frac{\eta_t}{2\lambda} + \frac{\mu}{4} + \frac{\tilde{L}}{2}-\frac{1}{\lambda}) \|\tilde{y}_{t+1}-y_t\|^2 \nonumber \\
  & \quad + \frac{2}{\mu} \|\nabla_y f(x_t,y_t)-w_t\|^2 \nonumber \\
  & \leq ( \frac{1}{2\eta_t\lambda}-\frac{\mu}{4})\|y_t -y^*(x_t)\|^2 + (\frac{3\tilde{L}}{4} -\frac{1}{2\lambda}) \|\tilde{y}_{t+1}-y_t\|^2 + \frac{2}{\mu} \|\nabla_y f(x_t,y_t)-w_t\|^2 \nonumber \\
  & = ( \frac{1}{2\eta_t\lambda}-\frac{\mu}{4})\|y_t -y^*(x_t)\|^2 - \big( \frac{3}{8\lambda} + \frac{1}{8\lambda} -\frac{3\tilde{L}}{4}\big) \|\tilde{y}_{t+1}-y_t\|^2 \nonumber \\
  &\quad + \frac{2}{\mu} \|\nabla_y f(x_t,y_t)-w_t\|^2 \nonumber \\
  & \leq  ( \frac{1}{2\eta_t\lambda}-\frac{\mu}{4})\|y_t -y^*(x_t)\|^2 - \frac{3}{8\lambda} \|\tilde{y}_{t+1}-y_t\|^2 + \frac{2}{\mu}\|\nabla_y f(x_t,y_t)-w_t\|^2,
 \end{align}
 where the second inequality holds by $\tilde{L} \geq L_{22} \geq \mu$ and $0< \eta_t\leq 1$, and the last inequality is due to
 $0< \lambda \leq \frac{1}{6\tilde{L}}$.
 It implies that
 \begin{align} \label{eq:E8}
 \|y_{t+1}-y^*(x_t)\|^2 \leq ( 1-\frac{\eta_t\mu\lambda}{2})\|y_t -y^*(x_t)\|^2 - \frac{3\eta_t}{4} \|\tilde{y}_{t+1}-y_t\|^2
 + \frac{4\eta_t\lambda}{\mu}\|\nabla_y f(x_t,y_t)-w_t\|^2.
 \end{align}

 Next, we decompose the term $\|y_{t+1} - y^*(x_{t+1})\|^2$ as follows:
 \begin{align} \label{eq:E9}
  \|y_{t+1} - y^*(x_{t+1})\|^2 & = \|y_{t+1} - y^*(x_t) + y^*(x_t) - y^*(x_{t+1})\|^2    \nonumber \\
  & =  \|y_{t+1} - y^*(x_t)\|^2 + 2\langle y_{t+1} - y^*(x_t), y^*(x_t) - y^*(x_{t+1})\rangle  + \|y^*(x_t) - y^*(x_{t+1})\|^2  \nonumber \\
  & \leq (1+\frac{\eta_t\mu\lambda}{4})\|y_{t+1} - y^*(x_t)\|^2  + (1+\frac{4}{\eta_t\mu\lambda})\|y^*(x_t) - y^*(x_{t+1})\|^2 \nonumber \\
  & \leq (1+\frac{\eta_t\mu\lambda}{4})\|y_{t+1} - y^*(x_t)\|^2  + (1+\frac{4}{\eta_t\mu\lambda})\eta^2_t\gamma^2\kappa^2\|v_t\|^2,
 \end{align}
 where the first inequality holds by the Cauchy-Schwarz inequality and Young's inequality,
 and the last equality is due to Lemma \ref{lem:A3}.

 By combining the above inequalities (\ref{eq:E8}) and (\ref{eq:E9}), we have
 \begin{align}
  \|y_{t+1} - y^*(x_{t+1})\|^2 & \leq (1+\frac{\eta_t\mu\lambda}{4})( 1-\frac{\eta_t\mu\lambda}{2})\|y_t -y^*(x_t)\|^2
  - (1+\frac{\eta_t\mu\lambda}{4})\frac{3\eta_t}{4} \|\tilde{y}_{t+1}-y_t\|^2     \nonumber \\
  & \quad + (1+\frac{\eta_t\mu\lambda}{4})\frac{4\eta_t\lambda}{\mu}\|\nabla_y f(x_t,y_t)-w_t\|^2
  + (1+\frac{4}{\eta_t\mu\lambda})\eta_t^2\gamma^2\kappa^2\|v_t\|^2.
 \end{align}
 Since $0 < \eta_t \leq 1$, $0< \lambda \leq \frac{1}{6\tilde{L}}$ and $\tilde{L}\geq L_{22}\geq \mu$,
 we have $\lambda \leq \frac{1}{6\tilde{L}} \leq \frac{1}{6\mu}$
 and $\eta_t \leq 1 \leq \frac{1}{6\mu\lambda }$. Then we obtain
 \begin{align}
  (1+\frac{\eta_t\mu\lambda}{4})( 1-\frac{\eta_t\mu\lambda}{2})&= 1-\frac{\eta_t\mu\lambda}{2} +\frac{\eta_t\mu\lambda}{4}
  - \frac{\eta_t^2\mu^2\lambda^2}{8} \leq 1-\frac{\eta_t\mu\lambda}{4}, \nonumber \\
  - (1+\frac{\eta_t\mu\lambda}{4})\frac{3\eta_t}{4} &\leq -\frac{3\eta_t}{4}, \nonumber \\
  (1+\frac{\eta_t\mu\lambda}{4})\frac{4\eta_t\lambda}{\mu} & \leq (1+\frac{1}{24})\frac{4\eta_t\lambda}{\mu}=\frac{25\eta_t\lambda}{6\mu}, \nonumber \\
  (1+\frac{4}{\eta_t\mu\lambda})\gamma^2\kappa^2\eta_t^2 & =  \gamma^2\kappa^2\eta_t^2 +\frac{4\gamma^2\kappa^2\eta_t}{\mu\lambda}
  \leq \frac{\gamma^2\kappa^2\eta_t}{6\mu\lambda} +\frac{4\gamma^2\kappa^2\eta_t}{\mu\lambda} = \frac{25\gamma^2\kappa^2\eta_t}{6\mu\lambda}.
 \end{align}
 Thus we have
 \begin{align}
      \|y_{t+1} - y^*(x_{t+1})\|^2 &\leq (1-\frac{\eta_t\mu\lambda}{4})\|y_t -y^*(x_t)\|^2 -\frac{3\eta_t}{4} \|\tilde{y}_{t+1}-y_t\|^2 \nonumber \\
      & \quad + \frac{25\eta_t\lambda}{6\mu}  \|\nabla_y f(x_t,y_t)-w_t\|^2 +  \frac{25\gamma^2\kappa^2\eta_t}{6\mu\lambda}\|v_t\|^2.
 \end{align}

\end{proof}

\subsection{ Convergence Analysis of RGDA and RSGDA Algorithms }
\label{Appendix:A1}
In the subsection, we study the convergence properties of our RGDA and RSGDA algorithms,
respectively.

\begin{theorem} \label{th:A1}
(Restatement of Theorem 1)
Suppose the sequence $\{x_t,y_t\}_{t=1}^T$ is generated from Algorithm \ref{alg:1} by using deterministic gradients.
Given $y_1=y^*(x_1)$, $\eta=\eta_t$ for all $t\geq 1$, $0< \eta \leq \min(1,\frac{1}{2\gamma L})$,
$0< \lambda \leq \frac{1}{6\tilde{L}}$ and $0< \gamma \leq \frac{\mu\lambda}{10\tilde{L}\kappa}$,
we have
\begin{align}
  \frac{1}{T} \sum_{t=1}^T  \|\mbox{grad} \ \Phi(x_t)\|
  \leq  \frac{2\sqrt{\Phi(x_1) - \Phi^*}}{\sqrt{\gamma \eta T}},
\end{align}
where $\tilde{L}=\max(1,L_{11},L_{12},L_{21},L_{22})$.
\end{theorem}

\begin{proof}
According to Lemma \ref{lem:A5}, we have
\begin{align} \label{eq:B1}
\|y_{t+1} - y^*(x_{t+1})\|^2 &\leq (1-\frac{\eta_t\mu\lambda}{4})\|y_t -y^*(x_t)\|^2 -\frac{3\eta_t}{4} \|\tilde{y}_{t+1}-y_t\|^2 + \frac{25\eta_t\lambda}{6\mu}  \|\nabla_y f(x_t,y_t)-w_t\|^2\nonumber \\
& \quad +  \frac{25\gamma^2\kappa^2\eta_t}{6\mu\lambda}\|v_t\|^2.
\end{align}
We first define a \emph{Lyapunov} function $\Lambda_t$, for any $t\geq 1$
\begin{align}
\Lambda_t = \Phi(x_t) + \frac{6\gamma\tilde{L}^2}{\lambda\mu}\|y_t - y^*(x_t)\|^2.
\end{align}
According to Lemma \ref{lem:A4}, we have
\begin{align}
\Lambda_{t+1} - \Lambda_t & = \Phi(x_{t+1}) - \Phi(x_t) + \frac{6\gamma\tilde{L}^2}{\lambda\mu}\big(\|y_{t+1} - y^*(x_{t+1})\|^2
-\|y_t - y^*(x_t)\|^2 \big) \nonumber \\
& \leq \gamma\eta_tL_{12} \|y_t - y^*(x_t)\|^2 + \gamma\eta_t\|\mbox{grad}_x f(x_t,y_t)-v_t\|^2 - \frac{\gamma\eta_t}{2}\|\mbox{grad} \Phi(x_t)\|^2 -\frac{\gamma\eta_t}{4}\|v_t\|^2 \nonumber \\
& \quad + \frac{6\gamma\tilde{L}^2}{\lambda\mu}\big( -\frac{\mu\lambda\eta_t}{4}\|y_t -y^*(x_t)\|^2 -\frac{3\eta_t}{4} \|\tilde{y}_{t+1}-y_t\|^2 + \frac{25\lambda\eta_t}{6\mu} \|\nabla_y f(x_t,y_t)-w_t\|^2 \nonumber \\
& \quad + \frac{25\gamma^2\kappa^2\eta_t}{6\mu\lambda}\|v_t\|^2\big) \nonumber \\
& \leq -\frac{\tilde{L}^2\gamma\eta_t}{2} \|y_t - y^*(x_t)\|^2- \frac{\gamma\eta_t}{2}\|\mbox{grad} \Phi(x_t)\|^2 - \frac{9\gamma\tilde{L}^2\eta_t}{2\lambda\mu}\|\tilde{y}_{t+1}-y_t\|^2 \nonumber \\
& \quad - \big( \frac{1}{4} - \frac{25\kappa^2\tilde{L}^2\gamma^2}{\mu^2\lambda^2}\big)\gamma\eta_t\|v_t\|^2 \nonumber \\
& \leq -\frac{\tilde{L}^2\gamma\eta_t}{2} \|y_t - y^*(x_t)\|^2- \frac{\gamma\eta_t}{2}\|\mbox{grad} \Phi(x_t)\|^2,
\end{align}
where the first inequality holds by the inequality (\ref{eq:B1});
the second last inequality is due to $\tilde{L}= \max(1,L_{11},L_{12},L_{21},L_{22})$ and $v_t=\mbox{grad}_xf(x_t,y_t)$, $w_t=\nabla_yf(x_t,y_t)$,
and the last inequality is due to $0< \gamma \leq \frac{\mu\lambda}{10\tilde{L}\kappa}$.
Thus, we obtain
\begin{align} \label{eq:B2}
\frac{\tilde{L}^2\gamma\eta_t}{2} \|y_t - y^*(x_t)\|^2+ \frac{\gamma\eta_t}{2}\|\mbox{grad} \Phi(x_t)\|^2 \leq \Lambda_t - \Lambda_{t+1}.
\end{align}

Since the initial solution satisfies $y_1=y^*(x_1)=\arg\max_{y \in \mathcal{Y}}f(x_1,y)$, we have
\begin{align} \label{eq:B3}
 \Lambda_1 = \Phi(x_1) + \frac{6\gamma\tilde{L}^2}{\lambda\mu} \|y_1-y^*(x_1)\|^2 = \Phi(x_1).
\end{align}
Taking average over $t=1,2,\cdots,T$ on both sides of the inequality (\ref{eq:B2}), we have
\begin{align}
  \frac{1}{T} \sum_{t=1}^T \big[\frac{\tilde{L}^2\eta_t}{2} \|y_t - y^*(x_t)\|^2 + \frac{\eta_t}{2}\|\mbox{grad} \Phi(x_t)\|^2  \big]
 & \leq \frac{\Lambda_1 - \Lambda_{T+1}}{\gamma T} \leq \frac{\Phi(x_1) - \Phi^*}{\gamma T},
\end{align}
where the last equality is due to the above equality (\ref{eq:B3}) and Assumption 4.
Let $\eta=\eta_1=\cdots=\eta_T$, we have
\begin{align}
 \frac{1}{T} \sum_{t=1}^T \big[\tilde{L}^2 \|y_t - y^*(x_t)\|^2 + \|\mbox{grad} \Phi(x_t)\|^2\big]
 \leq  \frac{2(\Phi(x_1) - \Phi^*)}{\gamma \eta T}.
\end{align}

According to Jensen's inequality, we have
\begin{align}
 \frac{1}{T} \sum_{t=1}^T\big[\tilde{L} \|y_t - y^*(x_t)\| + \|\mbox{grad} \Phi(x_t)\| \big] &\leq \bigg( \frac{2}{T} \sum_{t=1}^T
 \big[\tilde{L}^2 \|y_t - y^*(x_t)\|^2 + \|\mbox{grad} \Phi(x_t)\|^2\bigg)^{1/2} \nonumber \\
 & \leq \bigg( \frac{4(\Phi(x_1) - \Phi^*)}{\gamma \eta T}\bigg)^{1/2} = \frac{2\sqrt{\Phi(x_1) - \Phi^*}}{\sqrt{\gamma \eta T}}.
\end{align}
Since $\tilde{L}\|y_t - y^*(x_t)\|+ \|\mbox{grad} \Phi(x_t)\| \geq \|\mbox{grad} \Phi(x_t)\|$, we can obtain
\begin{align}
 \frac{1}{T} \sum_{t=1}^T \|\mbox{grad} \Phi(x_t)\| \leq \frac{2\sqrt{\Phi(x_1) - \Phi^*}}{\sqrt{\gamma \eta T}}.
\end{align}

\end{proof}

\begin{theorem} \label{th:A2}
(Restatement of Theorem 2)
Suppose the sequence $\{x_t,y_t\}_{t=1}^T$ is generated from Algorithm \ref{alg:1} by using stochastic gradients.
Given $y_1=y^*(x_1)$, $\eta=\eta_t$ for all $t\geq 1$, $0< \eta \leq \min(1,\frac{1}{2\gamma L})$,
$0< \lambda \leq \frac{1}{6\tilde{L}}$ and $0< \gamma \leq \frac{\mu\lambda}{10\tilde{L}\kappa}$,
we have
\begin{align}
  \frac{1}{T} \sum_{t=1}^T\mathbb{E} \|\mbox{grad} \ \Phi(x_t)\|
  \leq  \frac{2\sqrt{\Phi(x_1) - \Phi^*}}{\sqrt{\gamma \eta T}} + \frac{\sqrt{2}\sigma}{\sqrt{B}} + \frac{5\sqrt{2}\tilde{L}\sigma}{\sqrt{B}\mu}.
\end{align}
\end{theorem}

\begin{proof}
According to Lemma \ref{lem:A5}, we have
\begin{align} \label{eq:G1}
\|y_{t+1} - y^*(x_{t+1})\|^2 &\leq (1-\frac{\eta_t\mu\lambda}{4})\|y_t -y^*(x_t)\|^2 -\frac{3\eta_t}{4} \|\tilde{y}_{t+1}-y_t\|^2 + \frac{25\eta_t\lambda}{6\mu}  \|\nabla_y f(x_t,y_t)-w_t\|^2\nonumber \\
& \quad +  \frac{25\gamma^2\kappa^2\eta_t}{6\mu\lambda}\|v_t\|^2.
\end{align}
We first define a \emph{Lyapunov} function $\Theta_t$, for any $t\geq 1$
\begin{align}
\Theta_t = \mathbb{E}\big[\Phi(x_t) + \frac{6\gamma\tilde{L}^2}{\lambda\mu}\|y_t - y^*(x_t)\|^2 \big].
\end{align}
By Assumption 5, we have
\begin{align}
 & \mathbb{E}\|\mbox{grad}_x f(x_t,y_t)-v_t\|^2= \mathbb{E}\|\mbox{grad}_x f(x_t,y_t)-\frac{1}{B} \sum_{i=1}^B \mbox{grad}_x f(x_t,y_t;\xi^i_t)\|^2 \leq \frac{\sigma^2}{B},  \\
 & \mathbb{E}\|\nabla_y f(x_t,y_t)-w_t\|^2 = \mathbb{E}\|\nabla_y f(x_t,y_t)-\frac{1}{B} \sum_{i=1}^B \nabla_y f(x_t,y_t;\xi^i_t)\|^2\leq \frac{\sigma^2}{B}.
\end{align}

According to Lemma \ref{lem:A4}, we have
\begin{align}
\Theta_{t+1} - \Theta_t & = \mathbb{E}[\Phi(x_{t+1})] - \mathbb{E}[\Phi(x_t)] + \frac{6\gamma\tilde{L}^2}{\lambda\mu}\big(\mathbb{E}\|y_{t+1}
- y^*(x_{t+1})\|^2-\mathbb{E}\|y_t - y^*(x_t)\|^2 \big) \nonumber \\
& \leq \gamma\eta_t L_{12} \mathbb{E}\|y_t - y^*(x_t)\|^2 + \gamma\eta_t\mathbb{E}\|\mbox{grad}_x f(x_t,y_t)-v_t\|^2 - \frac{\gamma\eta_t}{2}\mathbb{E}\|\mbox{grad} \Phi(x_t)\|^2
-\frac{\gamma\eta_t}{4}\mathbb{E}\|v_t\|^2 \nonumber \\
& \quad + \frac{6\gamma\tilde{L}^2}{\lambda\mu}\big( -\frac{\mu\lambda\eta_t}{4} \mathbb{E}\|y_t -y^*(x_t)\|^2 -\frac{3\eta_t}{4} \mathbb{E}\|\tilde{y}_{t+1}-y_t\|^2 + \frac{25\lambda\eta_t}{6\mu} \mathbb{E}\|\nabla_y f(x_t,y_t)-w_t\|^2 \nonumber \\
& \quad + \frac{25\gamma^2\kappa^2\eta_t}{6\mu\lambda}\mathbb{E}\|v_t\|^2\big) \nonumber \\
& \leq -\frac{\tilde{L}^2\gamma\eta_t}{2} \mathbb{E}\|y_t - y^*(x_t)\|^2- \frac{\gamma\eta_t}{2}\mathbb{E}\|\mbox{grad} \Phi(x_t)\|^2 - \frac{9\gamma\tilde{L}^2\eta_t}{2\lambda\mu}\mathbb{E}\|\tilde{y}_{t+1}-y_t\|^2 \nonumber \\
& \quad - \big( \frac{1}{4} - \frac{25\kappa^2\tilde{L}^2\gamma^2}{\mu^2\lambda^2}\big)\gamma\eta_t\mathbb{E}\|v_t\|^2 + \gamma\eta_t\mathbb{E}\|\mbox{grad}_x f(x_t,y_t)-v_t\|^2
+ \frac{25\tilde{L}^2\gamma\eta_t}{\mu^2} \mathbb{E}\|\nabla_y f(x_t,y_t)-w_t\|^2 \nonumber \\
& \leq -\frac{\tilde{L}^2\gamma\eta_t}{2} \mathbb{E}\|y_t - y^*(x_t)\|^2- \frac{\gamma\eta_t}{2}\mathbb{E}\|\mbox{grad} \Phi(x_t)\|^2 + \frac{\gamma\eta_t\sigma^2}{B}
+ \frac{25\tilde{L}^2\gamma\eta_t\sigma^2}{B\mu^2},
\end{align}
where the first inequality holds by the inequality (\ref{eq:G1});
the second last inequality is due to $\tilde{L}= \max(1,L_{11},L_{12},L_{21},L_{22})$, and the last inequality is due to
$0< \gamma \leq \frac{\mu\lambda}{10\tilde{L}\kappa}$ and Assumption 5.
Thus, we obtain
\begin{align} \label{eq:G2}
\frac{\tilde{L}^2\gamma\eta_t}{2}\mathbb{E} \|y_t - y^*(x_t)\|^2 + \frac{\gamma\eta_t}{2}\mathbb{E}\|\mbox{grad} \Phi(x_t)\|^2 \leq \Theta_t - \Theta_{t+1} + \frac{\gamma\eta_t\sigma^2}{B} + \frac{25\tilde{L}^2\gamma\eta_t\sigma^2}{B\mu^2}.
\end{align}

Since the initial solution satisfies $y_1=y^*(x_1)=\arg\max_{y \in \mathcal{Y}}f(x_1,y)$, we have
\begin{align} \label{eq:G3}
 \Theta_1 = \Phi(x_1) + \frac{6\gamma\tilde{L}^2}{\lambda\mu} \|y_1-y^*(x_1)\|^2 = \Phi(x_1).
\end{align}
Taking average over $t=1,2,\cdots,T$ on both sides of the inequality (\ref{eq:G2}), we have
\begin{align}
  \frac{1}{T} \sum_{t=1}^T\mathbb{E} \big[\frac{\tilde{L}^2\eta_t}{2} \|y_t - y^*(x_t)\|^2 + \frac{\eta_t}{2}\|\mbox{grad} \Phi(x_t)\|^2  \big]
 & \leq \frac{\Theta_t - \Theta_{t+1}}{\gamma T} + \frac{1}{T}\sum_{t=1}^T\frac{\eta_t\sigma^2}{B} + \frac{1}{T}\sum_{t=1}^T\frac{25\tilde{L}^2\eta_t\sigma^2}{B\mu^2} \nonumber \\
 & = \frac{\Phi(x_1) - \Phi^*}{\gamma T} + \frac{1}{T}\sum_{t=1}^T\frac{\eta_t\sigma^2}{B} + \frac{1}{T}\sum_{t=1}^T\frac{25\tilde{L}^2\eta_t\sigma^2}{B\mu^2},
\end{align}
where the last equality is due to the above equality (\ref{eq:G3}).
Let $\eta=\eta_1=\cdots=\eta_T$, we have
\begin{align}
 \frac{1}{T} \sum_{t=1}^T\mathbb{E} \big[\tilde{L}^2 \|y_t - y^*(x_t)\|^2 + \|\mbox{grad} \Phi(x_t)\|^2\big]
 \leq  \frac{2(\Phi(x_1) - \Phi^*)}{\gamma \eta T} + \frac{\sigma^2}{B} + \frac{25\tilde{L}^2\sigma^2}{B\mu^2}.
\end{align}

According to Jensen's inequality, we have
\begin{align}
 \frac{1}{T} \sum_{t=1}^T\mathbb{E} \big[\tilde{L} \|y_t - y^*(x_t)\| + \|\mbox{grad} \Phi(x_t)\| \big] &\leq \big( \frac{2}{T} \sum_{t=1}^T\mathbb{E} \big[\tilde{L}^2 \|y_t - y^*(x_t)\|^2 + \|\mbox{grad} \Phi(x_t)\|^2\big)^{1/2} \nonumber \\
 & \leq \frac{4(\Phi(x_1) - \Phi^*)}{\gamma \eta T} + \frac{2\sigma^2}{B} + \frac{50\tilde{L}^2\sigma^2}{B\mu^2}\big)^{1/2} \nonumber \\
 & \leq \frac{2\sqrt{\Phi(x_1) - \Phi^*}}{\sqrt{\gamma \eta T}} + \frac{\sqrt{2}\sigma}{\sqrt{B}} + \frac{5\sqrt{2}\tilde{L}\sigma}{\sqrt{B}\mu},
\end{align}
where the last inequality is due to $(a_1+a_2+a_3)^{1/2} \leq a_1^{1/2} + a_2^{1/2} + a_3^{1/2}$ for all $a_1,a_2,a_3>0$.
Thus, we have
\begin{align}
 \frac{1}{T} \sum_{t=1}^T\mathbb{E} \|\mbox{grad} \Phi(x_t)\|\leq \frac{2\sqrt{\Phi(x_1) - \Phi^*}}{\sqrt{\gamma \eta T}} + \frac{\sqrt{2}\sigma}{\sqrt{B}} + \frac{5\sqrt{2}\tilde{L}\sigma}{\sqrt{B}\mu}.
\end{align}

\end{proof}

\subsection{Convergence Analysis of the Acc-RSGDA Algorithm}
\label{Appendix:A2}
In the subsection, we study the convergence properties of the Acc-RSGDA algorithm.

\begin{lemma} \label{lem:F1}
(Restatement of Lemma 4) 
 Suppose the stochastic gradients $v_t$ and $w_t$ is generated from Algorithm \ref{alg:2}, given
 $0<\alpha_{t+1}\leq 1$ and $0<\beta_{t+1}\leq 1$, we have
\begin{align} \label{eq:F1}
\mathbb{E} \|\mbox{grad}_x f(x_{t+1},y_{t+1}) - v_{t+1}\|^2  &\leq (1-\alpha_{t+1})^2 \mathbb{E} \|\mbox{grad}_x f(x_t,y_t) -v_t\|^2 + 4(1-\alpha_{t+1})^2L^2_{11}\gamma^2\eta^2_t\mathbb{E}\|v_t\|^2 \nonumber \\
& \quad + 4(1-\alpha_{t+1})^2L^2_{12}\eta^2_t\mathbb{E}\|\tilde{y}_{t+1}-y_t\|^2 + \frac{2\alpha_{t+1}^2\sigma^2}{B}.
\end{align}
\begin{align} \label{eq:F2}
 \mathbb{E} \|\nabla_y f(x_{t+1},y_{t+1}) - w_{t+1}\|^2 & \leq (1-\beta_{t+1})^2 \mathbb{E} \|\nabla_y f(x_t,y_t) -w_t\|^2
 + 4(1-\beta_{t+1})^2L^2_{21}\gamma^2\eta^2_t\mathbb{E}\|v_t\|^2 \nonumber \\
 & \quad + 4(1-\beta_{t+1})^2L^2_{22}\eta^2_t\mathbb{E}\|\tilde{y}_{t+1}-y_t\|^2 + \frac{2\beta_{t+1}^2\sigma^2}{B}.
\end{align}
\end{lemma}

\begin{proof}
We first prove the inequality (\ref{eq:F1}).
According to the definition of $v_t$ in Algorithm \ref{alg:2}, we have
\begin{align}
  v_{t+1}-\mathcal{T}^{x_{t+1}}_{x_t}v_t &= -\alpha_{t+1}\mathcal{T}^{x_{t+1}}_{x_t}v_t + (1-\alpha_{t+1})\big(\mbox{grad}_x f_{\mathcal{B}_{t+1}}(x_{t+1},y_{t+1}) - \mathcal{T}^{x_{t+1}}_{x_t}\mbox{grad}_x f_{\mathcal{B}_{t+1}}(x_t,y_t)\big) \nonumber \\
  & + \alpha_{t+1}\mbox{grad}_x f_{\mathcal{B}_{t+1}}(x_{t+1},y_{t+1}).
\end{align}
Then we have
 \begin{align} \label{eq:F3}
  & \mathbb{E} \|\mbox{grad}_x f(x_{t+1},y_{t+1}) - v_{t+1}\|^2  \\
  & = \mathbb{E} \|\mbox{grad}_x f(x_{t+1},y_{t+1}) - \mathcal{T}^{x_{t+1}}_{x_t}v_t - (v_{t+1}-\mathcal{T}^{x_{t+1}}_{x_t}v_t)\|^2 \nonumber \\
  & = \mathbb{E} \|\mbox{grad}_x f(x_{t+1},y_{t+1}) - \mathcal{T}^{x_{t+1}}_{x_t}v_t + \alpha_{t+1}\mathcal{T}^{x_{t+1}}_{x_t}v_t
  - \alpha_{t+1}\mbox{grad}_x f_{\mathcal{B}_{t+1}}(x_{t+1},y_{t+1})
  \nonumber \\
  & \quad  - (1-\alpha_{t+1})\big(\mbox{grad}_x f_{\mathcal{B}_{t+1}}(x_{t+1},y_{t+1})- \mathcal{T}^{x_{t+1}}_{x_t}\mbox{grad}_x f_{\mathcal{B}_{t+1}}(x_t,y_t)\big) \|^2
  \nonumber \\
  & = \mathbb{E} \|(1-\alpha_{t+1})\mathcal{T}^{x_{t+1}}_{x_t}(\mbox{grad}_x f(x_t,y_t) -v_t) + (1-\alpha_{t+1})\big(\mbox{grad}_x f(x_{t+1},y_{t+1})
  -\mathcal{T}^{x_{t+1}}_{x_t}\mbox{grad}_x f(x_t,y_t) \nonumber \\
  & \quad -\mbox{grad}_x f_{\mathcal{B}_{t+1}}(x_{t+1},y_{t+1}) + \mathcal{T}^{x_{t+1}}_{x_t}\mbox{grad}_x f_{\mathcal{B}_{t+1}}(x_t,y_t)\big) \nonumber \\
  &\quad + \alpha_{t+1}\big(\mbox{grad}_x f(x_{t+1},y_{t+1})- \mbox{grad}_x f_{\mathcal{B}_{t+1}}(x_{t+1},y_{t+1})\big)\|^2 \nonumber \\
  & = (1-\alpha_{t+1})^2\mathbb{E}\|\mbox{grad}_x f(x_t,y_t)-v_t\|^2 +
  \alpha_{t+1}^2 \mathbb{E} \|\mbox{grad}_x f(x_{t+1},y_{t+1})- \mbox{grad}_x f_{\mathcal{B}_{t+1}}(x_{t+1},y_{t+1})\|^2\nonumber \\
  & \quad +(1-\alpha_{t+1})^2\mathbb{E} \|\mbox{grad}_x f(x_{t+1},y_{t+1})
  - \mathcal{T}^{x_{t+1}}_{x_t}\mbox{grad}_x f(x_t,y_t) - \mbox{grad}_x f_{\mathcal{B}_{t+1}}(x_{t+1},y_{t+1})
  \nonumber \\
  & \quad+ \mathcal{T}^{x_{t+1}}_{x_t}\mbox{grad}_x f_{\mathcal{B}_{t+1}}(x_t,y_t)\|^2 + 2\alpha_{t+1}(1-\alpha_{t+1})\big\langle\mbox{grad}_x f(x_{t+1},y_{t+1})
  - \mathcal{T}^{x_{t+1}}_{x_t}\mbox{grad}_x f(x_t,y_t)  \nonumber \\
  & \quad - \mbox{grad}_x f_{\mathcal{B}_{t+1}}(x_{t+1},y_{t+1}) + \mathcal{T}^{x_{t+1}}_{x_t}\mbox{grad}_x f_{\mathcal{B}_{t+1}}(x_t,y_t),\mbox{grad}_x f(x_{t+1},y_{t+1})
  - \mbox{grad}_x f_{\mathcal{B}_{t+1}}(x_{t+1},y_{t+1})\big\rangle \nonumber \\
  & \leq (1-\alpha_{t+1})^2\mathbb{E} \|\mbox{grad}_x f(x_t,y_t) -v_t\|^2 + 2\alpha_{t+1}^2 \mathbb{E} \|\mbox{grad}_x f(x_{t+1},y_{t+1})- \mbox{grad}_x f_{\mathcal{B}_{t+1}}(x_{t+1},y_{t+1})\|^2  \nonumber \\
  & \quad + 2(1-\alpha_{t+1})^2\mathbb{E} \|\mbox{grad}_x f(x_{t+1},y_{t+1}) - \mathcal{T}^{x_{t+1}}_{x_t}\mbox{grad}_x f(x_t,y_t) - \mbox{grad}_x f_{\mathcal{B}_{t+1}}(x_{t+1},y_{t+1}) \nonumber \\
  & \quad + \mathcal{T}^{x_{t+1}}_{x_t}\mbox{grad}_x f_{\mathcal{B}_{t+1}}(x_t,y_t)\|^2 \nonumber \\
  & \leq (1-\alpha_{t+1})^2 \mathbb{E} \|\mbox{grad}_x f(x_t,y_t) -v_t\|^2 + \frac{2\alpha_{t+1}^2\sigma^2}{B}  \nonumber \\
  & \quad + 2(1-\alpha_{t+1})^2\underbrace{ \mathbb{E} \| \mbox{grad}_x f_{\mathcal{B}_{t+1}}(x_{t+1},y_{t+1}) - \mathcal{T}^{x_{t+1}}_{x_t}\mbox{grad}_x f_{\mathcal{B}_{t+1}}(x_t,y_t)\|^2}_{=T_1}, \nonumber
 \end{align}
where the fourth equality follows by $\mathbb{E}[\mbox{grad}_x f_{\mathcal{B}_{t+1}}(x_{t+1},y_{t+1})]=\mbox{grad}_x f(x_{t+1},y_{t+1})$ and $ \mathbb{E}[\mbox{grad}_x f_{\mathcal{B}_{t+1}}(x_{t+1},y_{t+1}) - \mbox{grad}_x f_{\mathcal{B}_{t+1}}(x_t,y_t)]=\mbox{grad}_x f(x_{t+1},y_{t+1}) - \mbox{grad}_x f(x_t,y_t)$; the first inequality holds by Young's inequality; the last inequality is due to the equality $\mathbb{E}\|\zeta-\mathbb{E}[\zeta]\|^2 =\mathbb{E}\|\zeta\|^2 - \|\mathbb{E}[\zeta]\|^2$ and Assumption 5.

Next, we consider an upper bound of the above term $T_1$ as follows:
 \begin{align}
 T_1 & = \mathbb{E} \big\| \mbox{grad}_x f_{\mathcal{B}_{t+1}}(x_{t+1},y_{t+1}) - \mathcal{T}^{x_{t+1}}_{x_t}\mbox{grad}_x f_{\mathcal{B}_{t+1}}(x_t,y_t)\big\|^2  \\
 & = \mathbb{E} \big\| \mbox{grad}_x f_{\mathcal{B}_{t+1}}(x_{t+1},y_{t+1}) -\mathcal{T}^{x_{t+1}}_{x_t}\mbox{grad}_x f(x_t,y_{t+1};\xi_{t+1}) + \mathcal{T}^{x_{t+1}}_{x_t}\mbox{grad}_x f(x_t,y_{t+1};\xi_{t+1}) \nonumber \\
 &\quad - \mathcal{T}^{x_{t+1}}_{x_t} \mbox{grad}_x f_{\mathcal{B}_{t+1}}(x_t,y_t)\big\|^2  \nonumber \\
 & \leq 2\mathbb{E} \big\| \mbox{grad}_x f_{\mathcal{B}_{t+1}}(x_{t+1},y_{t+1}) -\mathcal{T}^{x_{t+1}}_{x_t}\mbox{grad}_x f(x_t,y_{t+1};\xi_{t+1})\|^2 \nonumber \\
 & \quad + 2\mathbb{E} \|\mbox{grad}_x f(x_t,y_{t+1};\xi_{t+1}) - \mbox{grad}_x f_{\mathcal{B}_{t+1}}(x_t,y_t)\big\|^2  \nonumber \\
 & \leq 2L^2_{11}\gamma^2\eta^2_t\mathbb{E}\|v_t\|^2 + 2L^2_{12}\mathbb{E}\|y_{t+1}-y_t\|^2  \nonumber \\
 & = 2L^2_{11}\gamma^2\eta^2_t\mathbb{E}\|v_t\|^2 + 2L^2_{12}\eta^2_t\mathbb{E}\|\tilde{y}_{t+1}-y_t\|^2,
 \end{align}
where the last inequality is due to Assumption 1.
Thus, we have
\begin{align}
\mathbb{E} \|\mbox{grad}_x f(x_{t+1},y_{t+1}) - v_{t+1}\|^2  &\leq (1-\alpha_{t+1})^2 \mathbb{E} \|\mbox{grad}_x f(x_t,y_t) -v_t\|^2 + 4(1-\alpha_{t+1})^2L^2_{11}\gamma^2\eta^2_t\mathbb{E}\|v_t\|^2 \nonumber \\
& \quad + 4(1-\alpha_{t+1})^2L^2_{12}\eta^2_t\mathbb{E}\|\tilde{y}_{t+1}-y_t\|^2 + \frac{2\alpha_{t+1}^2\sigma^2}{B}.
\end{align}

We apply a similar analysis to prove the above inequality (\ref{eq:F2}). We obtain
\begin{align}
 \mathbb{E} \|\nabla_y f(x_{t+1},y_{t+1}) - w_{t+1}\|^2 & \leq (1-\beta_{t+1})^2 \mathbb{E} \|\nabla_y f(x_t,y_t) -w_t\|^2
 + 4(1-\beta_{t+1})^2L^2_{21}\gamma^2\eta^2_t\mathbb{E}\|v_t\|^2 \nonumber \\
 & \quad + 4(1-\beta_{t+1})^2L^2_{22}\eta^2_t\mathbb{E}\|\tilde{y}_{t+1}-y_t\|^2 + \frac{2\beta_{t+1}^2\sigma^2}{B}.
\end{align}

\end{proof}

\begin{theorem} \label{th:A4}
(Restatement of Theorem 3)
Suppose the sequence $\{x_t,y_t\}_{t=1}^T$ is generated from Algorithm \ref{alg:2}.
Given $y_1=y^*(x_1)$, $c_1 \geq \frac{2}{3b^3} + 2\lambda\mu$, $c_2 \geq \frac{2}{3b^3} + \frac{50\lambda\tilde{L}^2}{\mu}$, $b>0$, $m\geq \max\big( 2, (\tilde{c}b)^3\big)$, $0<\gamma \leq \frac{\mu\lambda}{2\kappa\tilde{L}\sqrt{25+4\mu\lambda}}$ and $0<\lambda\leq \frac{1}{6\tilde{L}}$, we have
\begin{align}
  \frac{1}{T} \sum_{t=1}^T \mathbb{E}\|\mbox{grad} \ \Phi(x_t)\|
  \leq \frac{\sqrt{2M'}m^{1/6}}{T^{1/2}} + \frac{\sqrt{2M'}}{T^{1/3}},
\end{align}
where $\tilde{c}=\max(2\gamma L,c_1,c_2,1)$ and
$M' = \frac{2(\Phi(x_1) - \Phi^*)}{\gamma b} + \frac{2\sigma^2}{\lambda\mu\eta_0bB} + \frac{2(c_1^2+c_2^2)\sigma^2 b^2}{\lambda\mu B}\ln(m+T)$.
\end{theorem}

\begin{proof}
Since $\eta_t$ is decreasing and $m\geq b^3$, we have $\eta_t \leq \eta_0 = \frac{b}{m^{1/3}} \leq 1$. Similarly, due to
$m\geq (2\gamma Lb)^3$, we have $\eta_t \leq \eta_0 = \frac{b}{m^{1/3}} \leq \frac{1}{2\gamma L}$.
Due to $0 < \eta_t \leq 1$ and $m\geq \max\big( (c_1b)^3, (c_2b)^3 \big)$, we have $\alpha_{t+1} = c_1\eta_t^2 \leq c_1\eta_t \leq \frac{c_1b}{m^{1/3}}\leq 1$ and $\beta_{t+1} = c_2\eta_t^2 \leq c_2\eta_t \leq \frac{c_2b}{m^{1/3}}\leq 1$.
According to Lemma \ref{lem:F1}, we have
 \begin{align}
  & \frac{1}{\eta_t}\mathbb{E} \|\mbox{grad}_x f(x_{t+1},y_{t+1}) - v_{t+1}\|^2 - \frac{1}{\eta_{t-1}}\mathbb{E} \|\mbox{grad}_x f(x_t,y_t) - v_t\|^2  \\
  & \leq \big(\frac{(1-\alpha_{t+1})^2}{\eta_t} - \frac{1}{\eta_{t-1}}\big)\mathbb{E} \|\mbox{grad}_x f(x_t,y_t) -v_t\|^2  + 4(1-\alpha_{t+1})^2L^2_{11}\gamma^2\eta_t\mathbb{E}\|v_t\|^2 \nonumber \\
  & \quad + 4(1-\alpha_{t+1})^2L^2_{12}\eta_t\mathbb{E}\|\tilde{y}_{t+1}-y_t\|^2 + \frac{2\alpha_{t+1}^2\sigma^2}{\eta_tB} \nonumber \\
  & \leq \big(\frac{1-\alpha_{t+1}}{\eta_t} - \frac{1}{\eta_{t-1}}\big)\mathbb{E} \|\mbox{grad}_x f(x_t,y_t) -v_t\|^2 + 4L^2_{11}\gamma^2\eta_t\mathbb{E}\|v_t\|^2
  + 4L^2_{12}\eta_t\mathbb{E}\|\tilde{y}_{t+1}-y_t\|^2 + \frac{2\alpha_{t+1}^2\sigma^2}{\eta_tB}\nonumber \\
  & = \big(\frac{1}{\eta_t} - \frac{1}{\eta_{t-1}} - c_1\eta_t\big)\mathbb{E} \|\mbox{grad}_x f(x_t,y_t) -v_t\|^2 + 4L^2_{11}\gamma^2\eta_t\mathbb{E}\|v_t\|^2
  + 4L^2_{12}\eta_t\mathbb{E}\|\tilde{y}_{t+1}-y_t\|^2 + \frac{2\alpha_{t+1}^2\sigma^2}{\eta_tB}, \nonumber
 \end{align}
where the second inequality is due to $0<\alpha_{t+1}\leq 1$.
By a similar way, we also obtain
 \begin{align}
  & \frac{1}{\eta_t}\mathbb{E} \|\nabla_y f(x_{t+1},y_{t+1}) - w_{t+1}\|^2
  -  \frac{1}{\eta_{t-1}}\mathbb{E} \|\nabla_y f(x_t,y_t) - w_t\|^2  \\
  & \leq \big(\frac{1}{\eta_t} - \frac{1}{\eta_{t-1}} - c_2\eta_t\big)\mathbb{E} \|\nabla_y f(x_t,y_t) -w_t\|^2 + 4L^2_{21}\gamma^2\eta_t\mathbb{E}\|v_t\|^2 + 4L^2_{22}\eta_t\mathbb{E}\|\tilde{y}_{t+1}-y_t\|^2 + \frac{2\beta_{t+1}^2\sigma^2}{\eta_tB}. \nonumber
 \end{align}
By $\eta_t = \frac{b}{(m+t)^{1/3}}$, we have
 \begin{align}
  \frac{1}{\eta_t} - \frac{1}{\eta_{t-1}} &= \frac{1}{b}\big( (m+t)^{\frac{1}{3}} - (m+t-1)^{\frac{1}{3}}\big) \nonumber \\
  & \leq \frac{1}{3b(m+t-1)^{2/3}} \leq \frac{1}{3b\big(m/2+t\big)^{2/3}} \nonumber \\
  & \leq \frac{2^{2/3}}{3b(m+t)^{2/3}} = \frac{2^{2/3}}{3b^3}\frac{b^2}{(m/2+t)^{2/3}}= \frac{2^{2/3}}{3b^3}\eta_t^2 \leq \frac{2}{3b^3}\eta_t,
 \end{align}
where the first inequality holds by the concavity of function $f(x)=x^{1/3}$, \emph{i.e.}, $(x+y)^{1/3}\leq x^{1/3} + \frac{y}{3x^{2/3}}$; the second inequality is due to $m\geq 2$, and
the last inequality is due to $0<\eta_t\leq 1$.
Let $c_1 \geq \frac{2}{3b^3} + 2\lambda\mu$, we have
 \begin{align} \label{eq:P1}
  & \frac{1}{\eta_t}\mathbb{E} \|\mbox{grad}_x f(x_{t+1},y_{t+1}) - v_{t+1}\|^2 - \frac{1}{\eta_{t-1}}\mathbb{E} \|\mbox{grad}_x f(x_t,y_t) - v_t\|^2  \\
  & \leq -2\lambda\mu\eta_t\mathbb{E} \|\mbox{grad}_x f(x_t,y_t) -v_t\|^2 + 4L^2_{11}\gamma^2\eta_t\mathbb{E}\|v_t\|^2
  + 4L^2_{12}\eta_t\mathbb{E}\|\tilde{y}_{t+1}-y_t\|^2 + \frac{2\alpha_{t+1}^2\sigma^2}{\eta_tB}. \nonumber
\end{align}
Let $c_2 \geq \frac{2}{3b^3} + \frac{50\lambda\tilde{L}^2}{\mu}$, we have
\begin{align} \label{eq:P2}
  & \frac{1}{\eta_t}\mathbb{E} \|\nabla_y f(x_{t+1},y_{t+1}) - w_{t+1}\|^2
  -  \frac{1}{\eta_{t-1}}\mathbb{E} \|\nabla_y f(x_t,y_t) - w_t\|^2  \\
  & \leq - \frac{50\lambda\tilde{L}^2}{\mu}\eta_t\mathbb{E} \|\nabla_y f(x_t,y_t) -w_t\|^2 + 4L^2_{21}\gamma^2\eta_t\mathbb{E}\|v_t\|^2 + 4L^2_{22}\eta_t\mathbb{E}\|\tilde{y}_{t+1}-y_t\|^2 + \frac{2\beta_{t+1}^2\sigma^2}{\eta_tB}. \nonumber
\end{align}
According to Lemma \ref{lem:A5}, we have
\begin{align} \label{eq:P3}
\|y_{t+1} - y^*(x_{t+1})\|^2 -\|y_t -y^*(x_t)\|^2 & \leq -\frac{\eta_t\mu\lambda}{4}\|y_t -y^*(x_t)\|^2
-\frac{3\eta_t}{4} \|\tilde{y}_{t+1}-y_t\|^2 \nonumber \\
& \quad + \frac{25\lambda\eta_t}{6\mu}  \|\nabla_y f(x_t,y_t)-w_t\|^2 + \frac{25\gamma^2\kappa^2\eta_t}{6\mu\lambda}\|v_t\|^2.
\end{align}

Next, we define a \emph{Lyapunov} function $\Omega_t$, for any $t\geq 1$
\begin{align}
 \Omega_t  =  \mathbb{E}\big[\Phi(x_t) + \frac{\gamma}{2\lambda\mu\eta_{t-1}}\big( \|\mbox{grad}_x f(x_t,y_t)-v_t\|^2 + \|\nabla_y f(x_t,y_t)-w_t\|^2 \big)  + \frac{6\gamma\tilde{L}^2}{\lambda\mu} \|y_t-y^*(x_t)\|^2\big] .
\end{align}
Then we have
\begin{align} \label{eq:P4}
 \Omega_{t+1} - \Omega_t 
 & = \mathbb{E}[\Phi(x_{t+1})] - \mathbb{E}[\Phi(x_t)] + \frac{6\gamma\tilde{L}^2}{\lambda\mu} \big( \mathbb{E}\|y_{t+1}-y^*(x_{t+1})\|^2 - \mathbb{E}\|y_t-y^*(x_t)\|^2 \big) \nonumber \\
 & \quad + \frac{\gamma}{2\lambda\mu}\big( \frac{1}{\eta_t}\mathbb{E}\|\mbox{grad}_x f(x_{t+1},y_{t+1})-v_{t+1}\|^2 - \frac{1}{\eta_{t-1}}\mathbb{E}\|\mbox{grad}_x f(x_t,y_t)-v_t\|^2  \nonumber \\
 & \quad + \frac{1}{\eta_t}\mathbb{E}\|\nabla_y f(x_{t+1},y_{t+1})-w_{t+1}\|^2- \frac{1}{\eta_{t-1}}\mathbb{E}\|\nabla_y f(x_t,y_t)-w_t\|^2 \big) \nonumber \\
 & \leq L_{12}\gamma\eta_t \mathbb{E}\|y_t - y^*(x_t)\|^2 + \gamma\eta_t\mathbb{E}\|\mbox{grad}_x f(x_t,y_t)-v_t\|^2 - \frac{\gamma\eta_t}{2}\mathbb{E}\|\mbox{grad} \Phi(x_t)\|^2
 -\frac{\gamma\eta_t}{4}\mathbb{E}\|v_t\|^2  \nonumber \\
 & \quad + \frac{6\gamma\tilde{L}^2}{\lambda\mu} \big( -\frac{\mu\lambda\eta_t}{4}\mathbb{E}\|y_t -y^*(x_t)\|^2
 -\frac{3\eta_t}{4}\mathbb{E} \|\tilde{y}_{t+1}-y_t\|^2 + \frac{25\lambda\eta_t}{6\mu} \mathbb{E}\|\nabla_y f(x_t,y_t)-w_t\|^2 +  \frac{25\gamma^2\kappa^2\eta_t}{6\mu\lambda}\mathbb{E}\|v_t\|^2 \big)  \nonumber \\
 & \quad  + \frac{\gamma}{2\lambda\mu}\big( -2\lambda\mu\eta_t\mathbb{E} \|\mbox{grad}_x f(x_t,y_t) -v_t\|^2 + 4L^2_{11}\gamma^2\eta_t\mathbb{E}\|v_t\|^2
  + 4L^2_{12}\eta_t\mathbb{E}\|\tilde{y}_{t+1}-y_t\|^2 + \frac{2\alpha_{t+1}^2\sigma^2}{\eta_tB}  \nonumber \\
 & \quad - \frac{50\lambda\tilde{L}^2}{\mu}\eta_t\mathbb{E} \|\nabla_y f(x_t,y_t) -w_t\|^2 + 4L^2_{21}\gamma^2\eta_t\mathbb{E}\|v_t\|^2 + 4L^2_{22}\eta_t\mathbb{E}\|\tilde{y}_{t+1}-y_t\|^2 + \frac{2\beta_{t+1}^2\sigma^2}{\eta_tB} \big) \nonumber \\
 & \leq -\frac{\gamma\tilde{L}^2\eta_t}{2}\mathbb{E}\|y_t - y^*(x_t)\|^2 - \frac{\gamma\eta_t}{2}\mathbb{E}\|\mbox{grad} \Phi(x_t)\|^2 -\frac{\gamma\tilde{L}^2\eta_t}{2\lambda\mu}\mathbb{E}\|\tilde{y}_{t+1}-y_t\|^2 -\big(\frac{\gamma}{4} - \frac{25\gamma^3\kappa^2\tilde{L}^2}{\mu^2\lambda^2} - \frac{4\gamma^3\tilde{L}^2}{\mu\lambda}\big)\eta_t\mathbb{E}\|v_t\|^2 \nonumber \\
 & \quad + \frac{\gamma\alpha_{t+1}^2\sigma^2}{\lambda\mu\eta_tB} + \frac{\gamma\beta_{t+1}^2\sigma^2}{\lambda\mu\eta_tB} \nonumber \\
 & \leq -\frac{\gamma\tilde{L}^2\eta_t}{2}\mathbb{E}\|y_t - y^*(x_t)\|^2 - \frac{\gamma\eta_t}{2}\mathbb{E}\|\mbox{grad} \Phi(x_t)\|^2
 + \frac{\gamma\alpha_{t+1}^2\sigma^2}{\lambda\mu\eta_tB} + \frac{\gamma\beta_{t+1}^2\sigma^2}{\lambda\mu\eta_tB},
 \end{align}
where the first inequality holds by Lemmas \ref{lem:A4} and the above inequalities (\ref{eq:P1}), (\ref{eq:P2}) and (\ref{eq:P3});
the second inequality is due to $\tilde{L}= \max(1,L_{11},L_{12},L_{21},L_{22})$;
the last inequality is due to $0\leq \gamma \leq \frac{\mu\lambda}{2\kappa\tilde{L}\sqrt{25+4\mu\lambda}}$ and $\kappa\geq 1$.

According to the above inequality (\ref{eq:P4}), we have
\begin{align}  \label{eq:P5}
  \frac{\gamma\eta_t}{2}\big( \mathbb{E}\|\mbox{grad} \Phi(x_t)\|^2 + \tilde{L}^2\mathbb{E}\|y_t - y^*(x_t)\|^2\big) \leq \Omega_t - \Omega_{t+1} + \frac{\gamma\alpha_{t+1}^2\sigma^2}{\lambda\mu\eta_tB} + \frac{\gamma\beta_{t+1}^2\sigma^2}{\lambda\mu\eta_tB}.
\end{align}
Taking average over $t=1,2,\cdots,T$ on both sides of the inequality (\ref{eq:P5}), we have
\begin{align}
  \frac{1}{T} \sum_{t=1}^T\eta_t \mathbb{E}\big( \|\mbox{grad} \Phi(x_t)\|^2 + \tilde{L}^2\|y_t - y^*(x_t)\|^2\big)
 & \leq  \sum_{t=1}^T \frac{2(\Omega_t - \Omega_{t+1})}{\gamma T} + \frac{1}{T}\sum_{t=1}^T\big( \frac{2\alpha_{t+1}^2\sigma^2}{\lambda\mu\eta_tB} + \frac{2\beta_{t+1}^2\sigma^2}{\lambda\mu\eta_tB}\big). \nonumber
\end{align}
Since the initial solution satisfies $y_1=y^*(x_1)=\arg\max_{y \in \mathcal{Y}}f(x_1,y)$, we have
\begin{align} \label{eq:P6}
 \Omega_1 &= \Phi(x_1) + \frac{6\gamma\tilde{L}^2}{\lambda\mu} \|y_1-y^*(x_1)\|^2 + \frac{\gamma}{2\lambda\mu}\big( \frac{1}{\eta_0}\mathbb{E}\|\mbox{grad}_x f(x_1,y_1)-v_1\|^2 + \frac{1}{\eta_0}\mathbb{E}\|\nabla_y f(x_1,y_1)-w_1\|^2 \big) \nonumber \\
 & = \Phi(x_1) + \frac{\gamma}{2\lambda\mu}\big( \frac{1}{\eta_0}\mathbb{E}\|\mbox{grad}_x f(x_1,y_1)-\mbox{grad}_x f_{\mathcal{B}_1}(x_1,y_1)\|^2 + \frac{1}{\eta_0}\mathbb{E}\|\nabla_y f(x_1,y_1)-\nabla_y f_{\mathcal{B}_1}(x_1,y_1)\|^2 \big) \nonumber \\
 & \leq \Phi(x_1) + \frac{\gamma\sigma^2}{\lambda\mu\eta_0B},
\end{align}
where the last inequality holds by Assumption 5.

Consider $\eta_t$ is decreasing, i.e., $\eta_T^{-1} \geq \eta_t^{-1}$ for any $0\leq t\leq T$, we have
 \begin{align}
 & \frac{1}{T} \sum_{t=1}^T \mathbb{E}\big( \|\mbox{grad} \Phi(x_t)\|^2 + \tilde{L}^2\|y_t - y^*(x_t)\|^2 \big)  \\
 &\leq \sum_{t=1}^T \frac{2(\Omega_t - \Omega_{t+1})}{T\gamma\eta_T} + \frac{1}{T\eta_T}\sum_{t=1}^T\big( \frac{2\alpha_{t+1}^2\sigma^2}{ \lambda\mu\eta_tB} + \frac{2\beta_{t+1}^2\sigma^2}{ \lambda\mu\eta_tB}\big) \nonumber \\
 & \leq \frac{1}{T\eta_T} \big( \frac{2\Phi(x_1)}{\gamma} + \frac{2\sigma^2}{ \lambda\mu\eta_0B}  - \frac{2\Phi^*}{\gamma} \big) + \frac{1}{T\eta_T}\sum_{t=1}^T\big( \frac{2\alpha_{t+1}^2\sigma^2}{ \lambda\mu\eta_tB} + \frac{2\beta_{t+1}^2\sigma^2}{ \lambda\mu\eta_tB}\big)
  \nonumber \\
 & = \frac{2(\Phi(x_1) - \Phi^*)}{T\gamma\eta_T} +\frac{2\sigma^2}{T \lambda\mu\eta_0\eta_TB} + \frac{2(c_1^2+c_2^2)\sigma^2}{T \eta_T\lambda\mu B}\sum_{t=1}^T\eta_t^3 \nonumber \\
 & \leq \frac{2(\Phi(x_1) - \Phi^*)}{T\gamma\eta_T} +\frac{2\sigma^2}{T \lambda\mu\eta_0\eta_TB} + \frac{2(c_1^2+c_2^2)\sigma^2}{T \eta_T\lambda\mu B}\int^T_1\frac{b^3}{m+t} dt\nonumber \\
 & \leq \frac{2(\Phi(x_1) - \Phi^*)}{T\gamma\eta_T} + \frac{2\sigma^2}{T \lambda\mu\eta_0\eta_TB} + \frac{2(c_1^2+c_2^2)\sigma^2b^3}{T \eta_T\lambda\mu  B}\ln(m+T) \nonumber \\
 & = \frac{2(\Phi(x_1) - \Phi^*)}{T\gamma b}(m+T)^{1/3} + \frac{2\sigma^2}{T \lambda\mu\eta_0 bB}(m+T)^{1/3}
 + \frac{2(c_1^2+c_2^2)\sigma^2 b^2}{T \lambda\mu B}\ln(m+T)(m+T)^{1/3}, \nonumber
\end{align}
where the third inequality holds by $\sum_{t=1}^T\eta^3_t \leq \int^T_1\eta^3_t dt$. Let $M' = \frac{2(\Phi(x_1) - \Phi^*)}{\gamma b} + \frac{2\sigma^2}{\lambda\mu\eta_0 b B} + \frac{2(c_1^2+c_2^2)\sigma^2 b^2}{\lambda\mu B}\ln(m+T)$,
we rewrite the above inequality as follows:
\begin{align}
 \frac{1}{T} \sum_{t=1}^T \mathbb{E}\big( \|\mbox{grad} \Phi(x_t)\|^2 + \tilde{L}^2\|y_t - y^*(x_t)\|^2 \big) \leq \frac{M'}{T}(m+T)^{1/3}.
\end{align}
According to Jensen's inequality, we have
\begin{align}
  \frac{1}{T} \sum_{t=1}^T \mathbb{E}\big( \|\mbox{grad} \Phi(x_t)\| + \tilde{L}\|y_t - y^*(x_t)\| \big)
 & \leq \bigg( \frac{2}{T} \sum_{t=1}^T \mathbb{E}\big( \|\mbox{grad} \Phi(x_t)\|^2 + \tilde{L}^2\|y_t - y^*(x_t)\|^2  \big)\bigg)^{1/2} \nonumber \\
 & \leq \frac{\sqrt{2M'}}{T^{1/2}}(m+T)^{1/6} \leq \frac{\sqrt{2M'}m^{1/6}}{T^{1/2}} + \frac{\sqrt{2M'}}{T^{1/3}},
\end{align}
where the last inequality is due to $(a_1+a_2)^{1/6} \leq a_1^{1/6} + a_2^{1/6}$ for all $a_1,a_2>0$.
Thus, we have
\begin{align}
  \frac{1}{T} \sum_{t=1}^T \mathbb{E}\|\mbox{grad} \Phi(x_t)\|\leq \frac{\sqrt{2M'}m^{1/6}}{T^{1/2}} + \frac{\sqrt{2M'}}{T^{1/3}}.
\end{align}

\end{proof}

\end{appendices}

\end{onecolumn}

\end{document}